\pdfoutput=1

\PassOptionsToPackage{table,xcdraw}{xcolor}

\documentclass[11pt,table]{article}

\usepackage[]{naacl2021}

\usepackage{times}
\usepackage{float}
\usepackage{latexsym}
\usepackage{stfloats}
\usepackage{amsmath,amssymb}
\usepackage{dsfont}
\usepackage{booktabs}
\usepackage{comment}
\usepackage{xspace}
\usepackage{todonotes}
\usepackage{enumitem}
\usepackage[normalem]{ulem}
\usepackage{multirow}
\usepackage{notation}
\usepackage{pgfplotstable}
\usepackage{longtable}
\usepackage{tabu}
\pgfplotsset{compat=1.16}

\usepackage[T1]{fontenc}

\usepackage[utf8]{inputenc}

\usepackage{microtype}

\title{Learning to Recognize Dialect Features}

\author{{\bf Dorottya Demszky}\textsuperscript{1}\thanks{\quad Work done while at Google Research.}\quad {\bf Devyani Sharma}\textsuperscript{2}\quad {\bf Jonathan H. Clark}\textsuperscript{3}\vspace{1mm}\\{\bf Vinodkumar Prabhakaran}\textsuperscript{3}\quad {\bf Jacob Eisenstein}\textsuperscript{3}\quad \vspace{1mm}\\\textsuperscript{1}Stanford Linguistics\quad \textsuperscript{2}Queen Mary University of London\quad\textsuperscript{3}Google Research\vspace{1mm}\\ \texttt{ddemszky@stanford.edu}\\\texttt{d.sharma@qmul.ac.uk}\\\texttt{\{jhclark,vinodkpg,jeisenstein\}@google.com}}

\begin{document}

\maketitle

\begin{abstract}
Building NLP systems that serve everyone requires accounting for dialect differences. But dialects are not monolithic entities: rather, distinctions between and within dialects are captured by the presence, absence, and frequency of dozens of dialect features in speech and text, such as the deletion of the copula in ``He $\varnothing$ running''. In this paper, we introduce the task of dialect feature detection, and present two multitask learning approaches, both based on pretrained transformers. 
 For most dialects, large-scale annotated corpora for these features are unavailable, making it difficult to train recognizers. We train our models on a small number of minimal pairs, building on how linguists typically define dialect features. Evaluation on a test set of 22 dialect features of Indian English demonstrates that these models learn to recognize many features with high accuracy, and that a few minimal pairs can be as effective for training as thousands of labeled examples. We also demonstrate the downstream applicability of dialect feature detection both as a measure of dialect density and as a dialect classifier.

\end{abstract}

\section{Introduction}
\begin{figure}
    \centering
\includegraphics[width=\linewidth]{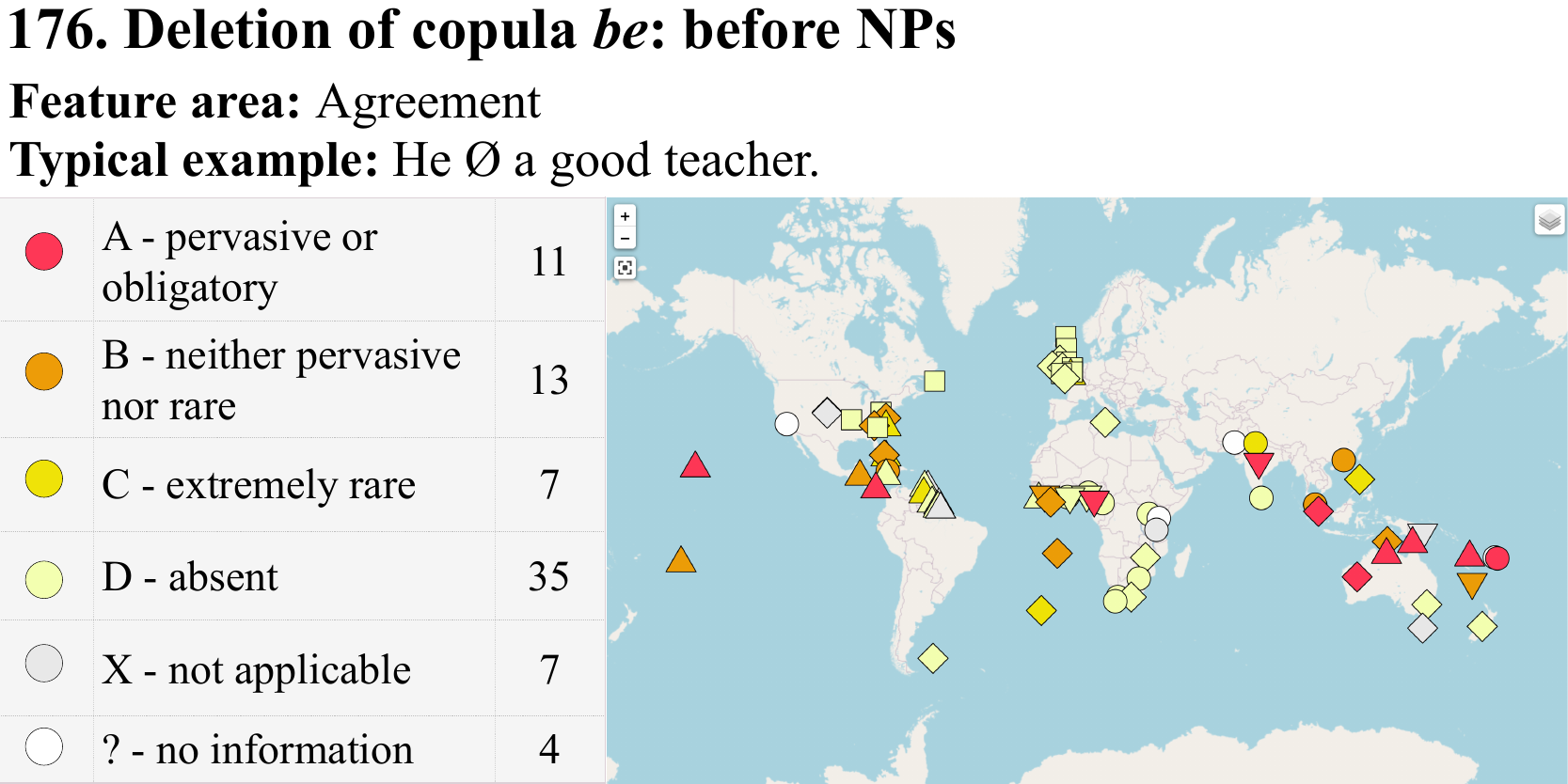}
\caption{An example dialect feature from the Electronic World Atlas of Varieties of English (eWAVE).\footnotemark}
    \label{fig:dialect-feature-example}
\end{figure}

Dialect variation is a pervasive property of language, 
which must be accounted for if we are to build robust natural language processing (NLP) systems that serve everyone. 
Linguists do not characterize dialects as simple categories, but rather as collections of correlated features~\cite{nerbonne2009data}, such as the one shown in \autoref{fig:dialect-feature-example}; speakers of any given dialect vary regarding which features they employ, how frequently, and in which contexts. 
In comparison to approaches that classify speakers or documents across dialects (typically using metadata such as geolocation), the feature-based perspective has several advantages: (1) allowing for fine-grained comparisons of speakers or documents \emph{within} dialects, without training on personal metadata; (2) disentangling grammatical constructions that make up the dialect from the content that may be frequently discussed in the dialect; (3) enabling robustness testing of NLP systems across dialect features, helping to ensure adequate performance even on cases other than ``high-resource'' varieties such as mainstream U.S. English~\cite{blodgett2016demographic}; (4) helping to develop more precise characterizations of dialects, enabling more accurate predictions of variable language use and better interpretations of its social implications~\cite[e.g.,][]{craig-washington-2002-ddm,vanhofwegen-wolfram-2010-ddm}.

\footnotetext{\url{https://ewave-atlas.org}. Shapes indicate variety type, e.g. creole, L1, and L2 English varieties.}
The main challenge for recognizing dialect features computationally is the lack of labeled data.   Annotating dialect features requires linguistic expertise and is prohibitively time-consuming given the large number of features and their sparsity. In dialectology, large-scale studies of text are limited to features that can be detected using regular expressions of surface forms and parts-of-speech, e.g., \regexp{\textsc{PRP} \textsc{DT}} for the copula deletion feature in \autoref{fig:dialect-feature-example}; many features cannot be detected with such patterns (e.g. \ofronting, \extraarticle). Furthermore, part-of-speech tagging is unreliable in many language varieties, such as regional and minority dialects~\cite{jorgensen2015challenges,blodgett2016demographic}. As dialect density correlates with social class and economic status~\cite{sahgal-1988-inde,Rickford-etal-2015-pnas,Grogger-etal-2020-wage}, the failure of language technology to cope with dialect differences may create allocational harms that reinforce social hierarchies~\cite{blodgett-etal-2020-language}. 

In this paper, we propose and evaluate learning-based approaches to recognize dialect features. We focus on Indian English, given the availability of domain expertise and labeled corpora for evaluation. 
First, we consider a standard multitask classification approach, in which a pretrained transformer~\cite{vaswani2017attention} 
is fine-tuned to recognize a set of dialect features. The architecture can be trained from two possible sources of supervision: (1) thousands of labeled corpus examples, (2) a small set of \emph{minimal pairs}, which are hand-crafted examples designed to highlight the key aspects of each dialect feature (as in the ``typical example'' field of \autoref{fig:dialect-feature-example}). Because most dialects have little or no labeled data,
the latter scenario is more realistic for most dialects. We also consider a multitask architecture that learns across multiple features by encoding the feature names, similar to recent work on few-shot or zero-shot multitask learning~\cite{logeswaran2019zero,brown2020language}.


In Sections~\ref{sec:evaluations} and \ref{sec:eval-density}, we discuss empirical evaluations of these models. Our main findings are:
\begin{itemize}[leftmargin=*,itemsep=0pt]
\item It is possible to detect individual dialect features: several features can be recognized with reasonably high accuracy. Our best models achieve a macro-AUC of $.848$ across ten grammatical features for which a large test set is available.
\item This performance can be obtained by training on roughly five minimal pairs per feature. Minimal pairs are significantly more effective for training than a comparable number of corpus examples.
\item Dialect feature recognizers can be used to rank documents by their density of dialect features, enabling within-dialect density computation for Indian English and accurate classification between Indian and U.S. English.
\end{itemize}



\begin{table*}
\small
\renewcommand{\arraystretch}{1.2}
    \centering
    \resizebox{.9\linewidth}{!}{%
    \begin{tabular}{llcc}
\toprule
\textbf{Feature} & \textbf{Example} & \multicolumn{2}{l}{\textbf{Count of Instantiations}}\\
\cmidrule{3-4}
& & \newcite{lange2012syntax} & Our data \\
\midrule
         \article{} &  \say{(the) chair is black} & & 59 \\
         \dodrop{} & \say{she doesn't like (it)} & & 14 \\
         \itself{} & \say{he is doing engineering in Delhi \uline{itself}} & 24 & 5 \\
         \only{} & \say{I was there yesterday \uline{only}} & 95 & 8 \\
         \habitual{} & \say{always we \uline{are giving} receipt} & & 2 \\
         \stative{} & \say{he \uline{is having} a television} & & 3 \\
         \whquestions{} & \say{what \uline{you are} doing?} & & 4 \\
         \agreement{} & \say{he \uline{do} a lot of things} & & 23 \\
         \leftdis & \say{\uline{my father}, he works for a solar company} & 300 & 19 \\
         \massnouns{} & \say{all the music\uline{s} \uline{are} very good} & & 13 \\
         \existential{} & \say{every year inflation \uline{is there}} &302 & 8 \\
         \ofronting{} & \say{\uline{minimum one month} you have to wait} &186 & 14 \\
         \ppfronting{} & \say{\uline{(on the) right side} we can see a plate} & & 11 \\
         \prepdrop{} & \say{I went (to) another school} & & 17 \\
         \embedded{} & \say{I don't know what {\uline{are they} doing}} & & 4 \\
         \invtag{} & \say{the children are outside, \uline{isn't it?}} & 786 & 17 \\
         \extraarticle{} & \say{she has \uline{a} business experience} & & 25 \\
         \andall{} & \say{then she did her schooling \uline{and all}} & & 7 \\
         \copula{} & \say{my parents (are) from Gujarat} & 71 \\
         \resobj{} & \say{my old life I want to spend \uline{it} in India} & 24 \\
         \ressubj{} & \say{my brother, \uline{he} lives in California} & 287 & \\
         \topnonarg{} & \say{\uline{in those years} I did not travel} &272 \\
         \bottomrule
    \end{tabular}%
    }
    \caption{Features of Indian English used in our evaluations and their counts in the two datasets we study.}
    \label{tab:features}
\end{table*}

\section{Data and Features of Indian English}
\label{sec:data}

We develop methods for detecting 22 dialect features associated with Indian English. Although India has over 125 million English speakers --- making it the world's second largest English-speaking population --- there is relatively little NLP research focused on Indian English. Our methods are not designed exclusively for specific properties of Indian English; many of the features that are associated with Indian English are also present in other dialects of English.

We use two sources of data in our study: an annotated corpus (\autoref{ssec:annotations}) and a dataset of minimal pairs (\autoref{ssec:minimal_pairs}). For evaluation, we use corpus annotations exclusively. The features are described in \autoref{tab:features}, and our data is summarized in \autoref{tab:data_stats}.

\begin{table}
    \centering
    \resizebox{\linewidth}{!}{%
\begin{tabular}{lccc}
\toprule
\multicolumn{2}{c}{\textbf{Dialect features}}   & \multicolumn{2}{c}{\textbf{Unique annotated examples}} \\
\cmidrule(lr){1-2}
\cmidrule(lr){3-4}
Feature set & Count & Corpus ex. & Min. pair ex.     \\
\midrule
Lange (2012)    & 10                    & 19059                     & 113                         \\
Extended            & 18                    & 367                       & 208     \\                         \bottomrule  
\end{tabular}%
}
    \caption{Summary of our labeled data. All corpus examples for the Lange features are from ICE-India; for the Extended feature set, examples are drawn from ICE-India and the Sharma data.}
    \label{tab:data_stats}
\end{table}

\subsection{Corpus Annotations}
\label{ssec:annotations}
The International Corpus of English~\cite[ICE;][]{greenbaum1996international}
is a collection of corpora of world varieties of English, organized primarily by the national origin of the speakers/writers. We focus on annotations of spoken dialogs (S1A-001 -- S1A-090) from the Indian English subcorpus (ICE-India). The ICE-India subcorpus was chosen in part because it is one of the only corpora with large-scale annotations of dialect features. To contrast Indian English with U.S. English (\autoref{sec:evaluations}), we use the Santa Barbara Corpus of Spoken American English~\cite{du2000santa} that constitutes the ICE-USA subcorpus of spoken dialogs.

We work with two main sources of dialect feature annotations in the ICE-India corpus:
\begin{description}
[itemindent=4pt,leftmargin=\labelsep-2pt]
\item[Lange features.]
The first set of annotations come from Claudia \newcite{lange2012syntax}, who annotated 10 features in 100 transcripts for an analysis of discourse-driven syntax in Indian English, such as topic marking and fronting. We use half of this data for training (50 transcripts, 9392 utterances), and half for testing (50 transcripts, 9667 utterances).
\item[Extended features.] To test a more diverse set of features, we additionally annotated 18 features on a set of 300 turns randomly selected from the conversational subcorpus of ICE-India,\footnote{We manually split turns that were longer than two clauses, resulting in 317 examples.} as well as 50 examples randomly selected from a secondary dataset of sociolinguistic interviews~\cite{sharma2009indian} to ensure diverse feature instantiation. We selected our 18 features based on multiple criteria: 1) prevalence in Indian English based on the dialectology literature, 2) coverage in the data (we started out with a larger set of features and removed those with fewer than two occurrences), 3) diversity of linguistic phenomena. The extended features overlap with those annotated by Lange, yielding a total set of 22 features.
Annotations were produced by consensus from the first two authors. To measure interrater agreement, a third author (JE) independently re-annotated 10\% of the examples, with Cohen's $\kappa=0.79$~\cite{cohen1960coefficient}.\footnote{Our annotations will be made available at \url{https://dialectfeatures.page.link/annotations}.}
\end{description}

\subsection{Minimal Pairs}
\label{ssec:minimal_pairs}
For each of the 22 features in \autoref{tab:features}, we created a small set of minimal pairs. The pairs were created by first designing a short example that demonstrated the feature, and then manipulating the example so that the feature is absent. This ``negative'' example captures the \emph{envelope of variation} for the feature, demonstrating a site at which the feature could be applied~\cite{labov1972sociolinguistic}. Consequently, negative examples in minimal pairs carry more information than in the typical annotation scenario, where absence of evidence does not usually imply evidence of absence. In our minimal pairs, the negative examples were chosen to be acceptable in standard U.S. and U.K. English, and can thus be viewed as situating dialects against standard varieties. Here are some example minimal pairs:
\begin{description}
[itemsep=3pt,parsep=0pt]
\item[\article{}:] \say{chair is black} $\to$ \say{\uline{the} chair is black}
\item[\only{}:] \say{I was there yesterday \uline{only}} $\to$ \say{I was there just yesterday}.
\item[\existential{}:] \say{every year inflation \uline{is there}} $\to$ \say{every year there is inflation}.
\end{description}

For most features, each minimal pair contains exactly one positive and one negative example. However, in some cases where more than two variants are available for an example  (e.g., for the feature \feature{\invtag{}}), we provide multiple positive examples to illustrate different variants. For Lange's set of 10 features, we provide a total of 113 unique examples; for the 18 extended features, we provide a set of 208 unique examples, roughly split equally between positives and negatives. The complete list of minimal pairs is included in Appendix~\ref{sec:appendix_minpairs}.
\section{Models and training}
\label{sec:models}
We train models to recognize dialect features by fine-tuning the BERT-base uncased transformer architecture~\cite{devlin2019bert}. We consider two strategies for constructing training data, and two architectures for learning across multiple features.

\subsection{Sources of supervision}
We consider two possible sources of supervision:

\begin{figure}

\begin{tabular}{|l|p{2.5in}|}
\hline
$y$ & $x$ \\
\hline
$1$ & \clstok article omission \septok Chair is black. \septok \\
$0$ & \clstok article omission \septok The chair is black. \septok \\
$0$ & \clstok article omission \septok I was there yesterday only. \septok \\
\ldots & \ldots \\
$1$ & \clstok focus only \septok I was there yesterday only. \septok\\
$0$ & \clstok focus only \septok I was there just yesterday. \septok\\
$0$ & \clstok focus only \septok Chair is black. \septok \\
\ldots &\ldots \\
\hline
    \end{tabular}
    \caption{Conversion of minimal pairs to labeled examples for \damodel, using two minimal pairs.}
    \label{fig:labeled-data-creation}
\end{figure}

\begin{description}[leftmargin=0pt]
\item[Minimal pairs.] We apply a simple procedure to convert minimal pairs into training data for classification. The positive part of each pair is treated as a positive instance for the associated feature, and the negative part is treated as a negative instance. Then, to generate more data, we also include elements of other minimal pairs as examples for each feature: for instance, a positive example of the \feature{\resobj{}} feature would be a negative example for \feature{\only{}}, unless the example happened to contain both features (this was checked manually). In this way, we convert the minimal pairs into roughly 113 examples per feature for Lange's features and roughly 208 examples per feature for the extended features. The total number of unique surface forms is still 113 and 208 respectively. Given the lack of labeled data for most dialects of the world, having existing minimal pairs or collecting a small number of minimal pairs is the most realistic data scenario.
\item[Corpus annotations.]
When sufficiently dense annotations are available, we can train a classifier based on these labeled instances. We use 50 of the ICE-India transcripts annotated by Lange, which consists of 9392 labeled examples (utterances) per feature. While we are lucky to have such a large resource for the Indian English dialect, this high-resource data scenario is rare.
\end{description}

\subsection{Architectures}
\label{sec:classification-architectures}

We consider two classification architectures:
\begin{description}
[leftmargin=0pt]
\item[Multihead.] In this architecture, which is standard for multitask classification, we estimate a linear \emph{prediction head} for each feature, which is simply a vector of weights. This is a multitask architecture, because the vast majority of model parameters from the input through the deep BERT stack remain shared among dialect features. The prediction head is then multiplied by the BERT embedding for the \clstok token to obtain a score for a feature's applicability to a given instance.
\item[\damodel.] Due to the few-shot nature of our prediction task, we also consider an architecture that attempts to exploit the natural language descriptions of each feature. This is done by concatenating the feature description to each element of the minimal pair. The instance is then labeled for whether the feature is present. This construction is shown in \autoref{fig:labeled-data-creation}. Prediction is performed by learning a single linear prediction head on the \clstok token. We call this model \emph{description-aware multitask learning}, or \damodel. 
\end{description}

\paragraph{Model details.} Both architectures are built on top of the BERT-base uncased model, which we fine-tune by cross-entropy for $500$ epochs (due to the small size of the training data) using the Adam optimizer~\cite{kingma2014adam}, batch size of $32$ and a learning rate of $10^{-5}$, warmed up over the first $150$ epochs. Annotations of dialect features were not used for hyperparameter selection. Instead, the hyperparameters were selected to maximize the discriminability between corpora of Indian and U.S. English, as described in \autoref{sec:eval-density-classification}. 
All models trained in less than two hours on a pod of four v2 TPU chips, with the exception of \damodel on corpus examples, which required up to 18 hours. 


\subsection{Regular Expressions}
In dialectology, regular expression pattern matching is the standard tool for recognizing dialect features~\cite[e.g.,][]{nerbonne2011gabmap}.
For the features described in \autoref{tab:features}, we were able to design regular expressions for only five.\footnote{Features: \feature{\itself{}, \only{}, \existential{}, \invtag{}}, and \feature{\andall{}}. \autoref{tab:regexes} lists all regular expressions.} Prior work sometimes relies on patterns that include both surface forms and part-of-speech~\cite[e.g.,][]{bohmann2019variation}, but part-of-speech cannot necessarily be labeled automatically for non-standard dialects~\cite{jorgensen2015challenges,blodgett2016demographic}, so we consider only regular expressions over surface forms.

\section{Results on Dialect Feature Detection}
\label{sec:evaluations}

In this section, we present results on the detection of individual dialect features. Using the features shown in \autoref{tab:features}, we compare supervision sources (corpus examples versus minimal pairs) and classification architectures (multihead versus \damodel) as described in \autoref{sec:models}. To avoid tuning a threshold for detection, we report area under the ROC curve (ROC-AUC), which has a value of $.5$ for random guessing and $1$ for perfect prediction.\footnote{Results for area under the precision-recall (AUPR) curve are shown in \autoref{app:avg-prec}. According to this metric, minimal pairs are less effective than the full training set of corpus examples, on average.}

\begin{table}
\centering
\pgfkeys{/pgf/number format/.cd,fixed,fixed zerofill,precision=3}
\setlength{\tabcolsep}{0.5ex}
\resizebox{\linewidth}{!}{%
\begin{filecontents*}{tables/lange_ROC_AUC.csv}
shortnames,finetune_annotated_data_lange,finetune_annotated_data_multilabel_lange,finetune_minimal_pairs_lange,finetune_minimal_pairs_multilabel_lange
\itself*,0.944825580468837,0.9253987984255231,0.9743741135599434,0.9596866982199488
\only*,0.9748164037508193,0.9106970573573916,0.9940369979314758,0.9384168527815173
\textsc{invariant tag},0.9907692331812552,0.9847364971857706,0.9685350328457426,0.9251072229277393
\copula,0.5362889812889813,0.640569956208254,0.6255226257353916,0.7456955810147299
\leftdis,0.8552240826635792,0.8793125282932657,0.7647201134974214,0.8850919698600039
\existential*,0.9905966202241568,0.9917605865544541,0.9053188511323246,0.8791562621448129
\ofronting,0.8054354201697101,0.8085091698275525,0.6783756305478434,0.7605606499041959
\textsc{res. object pronoun},0.5950633354056574,0.6665183659724381,0.7331299865298933,0.8253393430732567
\textsc{res. subject pronoun},0.8858497035435431,0.8868664388119196,0.6884511931061154,0.8573357903652825
\textsc{topicalized non-arg. const.},0.7245723706542736,0.7272675233140156,0.4988492658630969,0.7069749177606202
\textbf{Macro Average},0.8303441731350812,0.8421636921950586,0.7831313810749249,0.8483365288052107
\end{filecontents*}
\pgfplotstabletypeset[
    header=true,
    col sep=comma,
    white space chars={_},
    column type={lcccc},
    every head row/.style={
    before row={\toprule {\bf Supervision:} & \multicolumn{2}{c}{\bf Corpus examples} & \multicolumn{2}{c}{\bf Minimal pairs}\\},
    after row=\midrule},
    every last row/.style={after row=\bottomrule},    
    columns/shortnames/.style={string type,column name=Dialect feature},
    columns/finetune annotated data multilabel lange/.style={column name=Multihead},
    columns/finetune annotated data lange/.style={column name=\damodel},
    columns/finetune minimal pairs multilabel lange/.style={column name=Multihead},
    columns/finetune minimal pairs lange/.style={column name=\damodel}]
    {tables/lange_ROC_AUC.csv}
}
\caption{ROC-AUC scores on the Lange feature set, averaged across five random seeds. Asterisk (*) marks features that can be detected with relatively high accuracy ($>0.85$ ROC-AUC) using regular expressions.}
\label{tab:results-lange-features}
\end{table}

\subsection{Results on Lange Data and Features}

\begin{table*}
\resizebox{\linewidth}{!}{%
\begin{tabular}{@{}rllc|ccc}

\toprule
\multicolumn{1}{l}{}                   &                                                                                                                                           &                            &                                     &                                  & \multicolumn{2}{c}{\textbf{Multihead}}                \\
\multicolumn{1}{l}{\multirow{-2}{*}{}} & \multirow{-2}{*}{\textbf{Example}}                                                                                                                 & \multirow{-2}{*}{\textbf{Feature}}  & \multirow{-2}{*}{\textbf{\begin{tabular}[c]{@{}c@{}}Gold\\ label\end{tabular}}} & \multirow{-2}{*}{\textbf{Regex}} & Corpus ex.           & Min. pair       \\
\midrule 
1                                      & \begin{tabular}[c]{@{}l@{}}But whereas in Hyderabad they are only stuck with their books\\ and home and work that's all like\end{tabular} & \only               & 0                                   & \cellcolor[HTML]{EDA247}1        & \cellcolor[HTML]{EDA247}1 & \cellcolor[HTML]{57C4AD}0 \\[10pt] 
2                                      & \begin{tabular}[c]{@{}l@{}}There is there is a club this humour club oh good and I've chance\\ I had a chance of attending\end{tabular}   & \existential    & 0                                   & \cellcolor[HTML]{EDA247}1        & \cellcolor[HTML]{57C4AD}0 & \cellcolor[HTML]{57C4AD}0 \\[10pt]  
3                                      & New Education Policy is there isn't it?                                                                                                  & \existential    & 1                                   & \cellcolor[HTML]{57C4AD}1        & \cellcolor[HTML]{EDA247}0 & \cellcolor[HTML]{EDA247}0 \\[10pt] 
4                                      & I didn't go anywhere no                                                                                                                   & \invtag              & 0                                   & \cellcolor[HTML]{EDA247}1        & \cellcolor[HTML]{EDA247}1 & \cellcolor[HTML]{EDA247}1 \\[10pt] 
5                                      & \begin{tabular}[c]{@{}l@{}}In fact my son and daughter they had asked me to buy buy\\ them this thing the sunglasses\end{tabular}         & \leftdis           & 1                                   & \cellcolor[HTML]{EFEFEF}N/A      & \cellcolor[HTML]{57C4AD}1 & \cellcolor[HTML]{57C4AD}1 \\[10pt] 
6                                      & His house he is going to college KK diploma electronics                                                                                   & \ressubj & 0                                   & \cellcolor[HTML]{EFEFEF}N/A      & \cellcolor[HTML]{57C4AD}0 & \cellcolor[HTML]{EDA247}1 \\[10pt] 
7                                      & Which October first I think                                                                                                               & \copula           & 0                                   & \cellcolor[HTML]{EFEFEF}N/A      & \cellcolor[HTML]{EDA247}1 & \cellcolor[HTML]{57C4AD}0 \\[10pt] 
8                                      & Papers we can't say hard only because they already taught that same                                                                       & \copula             & 1                                   & \cellcolor[HTML]{EFEFEF}N/A      & \cellcolor[HTML]{EDA247}0 & \cellcolor[HTML]{EDA247}0 \\[10pt] 
9                                      & Just typing work I have to do                                                                                                             & \ofronting           & 1                                   & \cellcolor[HTML]{EFEFEF}N/A      & \cellcolor[HTML]{57C4AD}1 & \cellcolor[HTML]{57C4AD}1 \\[10pt]
10                                     & My post graduation degree I finished it in mid June nineteen eighty-six                                                                   & \resobj  & 1                                   & \cellcolor[HTML]{F3F3F3}N/A      & \cellcolor[HTML]{EDA247}0 & \cellcolor[HTML]{57C4AD}1 \\ \bottomrule
\end{tabular}
}
\caption{Example model predictions from the Lange data and feature set, comparing regular expressions with two versions of the multihead model, one trained on corpus examples and another on minimal pairs. `Gold label' indicates whether the feature was manually labeled as present in the original Lange data. Green and red indicate correct and incorrect predictions, respectively.}
    \label{tab:error_analysis}
\end{table*}

We first consider the 10 syntactic features from \newcite{lange2012syntax}, for which we have large-scale annotated data: the 100 annotated transcripts from the ICE-India corpus are split 50/50 into training and test sets. 
As shown in \autoref{tab:results-lange-features}, it is possible to achieve a Macro-AUC approaching .85 overall with multihead predictions on minimal pair examples. 
This is promising, because it suggests the possibility of recognizing dialect features for which we lack labeled corpus examples -- and such low-data situations are by far the most common data scenario among the dialects of the world.

The multihead architecture outperforms \damodel on both corpus examples and minimal pairs. In an ablation, we replaced the feature descriptions with non-descriptive identifiers such as ``Feature 3''. This reduced the Macro-AUC from to $.80$ with corpus examples, and to $.76$ with minimal pairs (averaged over five random seeds). We also tried longer feature descriptions, but this did not improve performance.

Unsurprisingly, the lexical features (e.g., \feature{\itself{}}) are easiest to recognize. The more syntactical features (e.g., \feature{\copula}, \feature{\resobj}) are more difficult, although some movement-based features (e.g., \feature{\leftdis}, \feature{\ressubj}) can be recognized accurately.

\paragraph{Qualitative model comparison.} 
We conducted a qualitative comparison of three models: regular expressions and two versions of the multihead model, one trained on corpus examples and another trained on minimal pairs. Table~\ref{tab:error_analysis} includes illustrative examples for the Lange data and features where models make different predictions. We find that the minimal pair model is better able to account for rare cases (e.g. use of non-focus ``only'' in Example 1), likely as it was trained on a few carefully selected set of examples illustrating positives and negatives. Both multihead models are able to account for disfluencies and restarts, in contrast to regular expressions (Example 2). Our analysis shows that several model errors are accounted for by difficult examples (Example 3: ``is there'' followed by ``isn't''; Example 6: restart mistaken for left dislocation) or the lack of contextual information available to the model (Example 4 \& 7: truncated examples). Please see Appendix~\ref{sec:appendix_outputs} for more details and random samples of model predictions.

\begin{figure}
    \centering
\includegraphics[width=\linewidth]{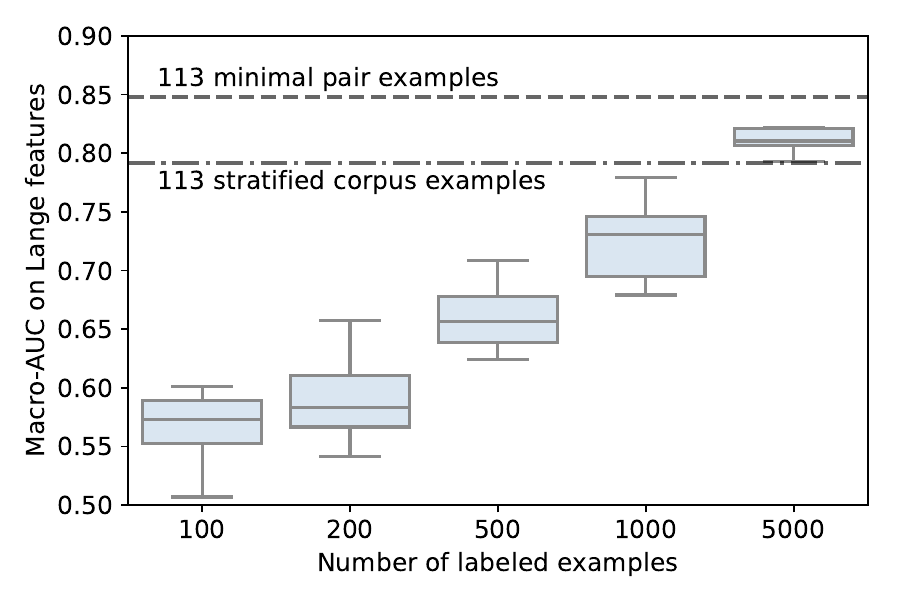}
\caption{Performance of the multihead model as the number of corpus examples is varied. Box plots are over 10 random data subsets, showing the 25th, 50th, and 75th percentiles; whiskers show the most extreme points within $\pm 1.5$ times the inter-quartile range.
}
    \label{fig:performance-by-number-of-examples}
\end{figure}

\paragraph{Learning from fewer corpus examples.} The minimal pair annotations consist of 113 examples; in contrast, there are 9392 labeled corpus examples, requiring far more effort to create. We now consider the situation when the amount of labeled data is reduced, focusing on the Lange features (for which labeled training data is available). As shown in \autoref{fig:performance-by-number-of-examples}, even 5000 labeled corpus examples do not match the performance of training on roughly 5 minimal pairs per feature.

%
\paragraph{Corpus examples stratified by feature.}
One reason that subsampled datasets yield weaker results is that they lack examples for many features. To enable a more direct comparison of corpus examples and minimal pairs, we created a set of ``stratified'' datasets of corpus examples, such that the number of positive and negative examples for each feature exactly matches the minimal pair data.
Averaged over ten such random stratified samples, the multihead model achieves a Macro-AUC of $.790$ ($\sigma = 0.029$), and \damodel achieves a Macro-AUC of $.722$ ($\sigma = .020$). These results are considerably worse than training on an equivalent number of minimal pairs, where the multihead model achieves a Macro-AUC of $.848$ and \damodel achieves a Macro-AUC of $.783$. This demonstrates the utility of minimal pairs over corpus examples for learning to recognize dialect features.

\subsection{Results on Extended Feature Set} Next, we consider the extended features, for which we have sufficient annotations for testing but not training (\autoref{tab:features}). Here we compare the \damodel and multihead models, using minimal pair data in both cases. As shown in \autoref{tab:results-our-features}, performance on these features is somewhat lower than on the Lange features, and for several features, at least one of the recognizers does worse than chance: 
\feature{\dodrop}, \feature{\extraarticle},
\feature{\massnouns}. These features seem to require deeper syntactic and semantic analysis, which may be difficult to learn from a small number of minimal pairs. On the other extreme, features with a strong lexical signature are recognized with high accuracy: \feature{\andall}, \feature{\itself}, \feature{\only}. These three features can also be recognized by regular expressions, as can \feature{\existential{}}.\footnote{\regexp{\textbackslash band all\textbackslash b}, \regexp{\textbackslash bitself\textbackslash b}, \regexp{\textbackslash bonly\textbackslash b}, \regexp{\textbackslash bis there\textbackslash b|\textbackslash bare there\textbackslash b}} However, for a number of other features, it is possible to learn a fairly accurate recognizer from just five minimal pairs: \feature{\article}, \feature{\embedded}, \feature{\leftdis}, \feature{\whquestions}. 

\subsection{Summary of Dialect Feature Detection}
Many dialect features can be automatically recognized with reasonably high discriminative power, as measured by area under the ROC curve. 
However, there are also features that are difficult to recognize: particularly, features of omission (such as \feature{\dodrop} and \feature{\prepdrop}), and the more semantic features such as \feature{\massnouns}. While some features can also be identified through regular expressions (e.g., \feature{\only}), there are many features that can be learned but cannot be recognized by regular expressions. We now move from individual features to aggregate measures of dialect density.

\begin{table}
\centering
\pgfkeys{/pgf/number format/.cd,fixed,fixed zerofill,precision=3}
\setlength{\tabcolsep}{0.5ex}
\resizebox{\linewidth}{!}{%
\begin{filecontents*}{tables/ours_ROC_AUC.csv}
shortnames,finetune_minimal_pairs_multilabel_ours,finetune_minimal_pairs_ours
\article,0.6576790814244914,0.5806726117619629
\dodrop,0.5629568106312293,0.49289867109634555
\extraarticle,0.4653942115768463,0.5455214570858283
\itself*,0.9488951841359773,1.0
\only*,0.7751428571428571,0.9980357142857142
\habitual,0.7176966292134832,0.43904494382022474
\textsc{invariant tag},0.9009487666034156,0.9838882180438159
\embedded,0.8844632768361581,0.7194915254237289
\agreement,0.6737600678253497,0.5429701851066836
\whquestions,0.6603107344632769,0.6491525423728813
\leftdis,0.819872347766086,0.7579092634121097
\massnouns,0.4654180602006689,0.44274247491638796
\existential*,0.8853571428571427,0.8965714285714286
\ofronting,0.7892441860465117,0.7217607973421927
\prepdrop,0.6476108331895808,0.5000862515094013
\ppfronting,0.6965679853287922,0.6552266177626408
\stative,0.7885446009389672,0.6448826291079813
\andall,0.9906389906389907,0.9938949938949939
\textbf{Macro Average},0.7405834314899903,0.6980416847507956
\end{filecontents*}
\pgfplotstabletypeset[
    header=true,
    col sep=comma,
    white space chars={_},
    column type={lcc},
    columns={shortnames,finetune minimal pairs ours,finetune minimal pairs multilabel ours},
    columns/shortnames/.style={string type,column name={\bf Dialect feature}},
        columns/finetune minimal pairs ours/.style={column name={\bf \damodel}},
    columns/finetune minimal pairs multilabel ours/.style={column name={\bf Multihead}},
    every head row/.style={before row={\toprule}, after row=\midrule},
    every last row/.style={after row=\bottomrule},    ]{tables/ours_ROC_AUC.csv}
}
\caption{ROC-AUC results on the extended feature set, averaged across five random seeds. Because labeled corpus examples are not available for some features, we train only on minimal pairs. Asterisk (*) marks features that can be detected with relatively high accuracy ($>0.85$ ROC-AUC) using regular expressions.}
\label{tab:results-our-features}
\end{table}

\section{Measuring Dialect Density}
\label{sec:eval-density}
A dialect density measure (DDM) is an aggregate over multiple dialect features that tracks the vernacularity of a passage of speech or text. Such measures are frequently used in dialectology~\cite{vanhofwegen-wolfram-2010-ddm}, and are also useful in research on education~\cite[e.g.,][]{craig-washington-2002-ddm}. Recently, a DDM was used to evaluate the performance of speech recognition systems by the density of AAVE features~\cite{koenecke2020racial}. The use of DDMs reflects the reality that speakers construct individual styles drawing on linguistic repertoires such as dialects to varying degrees~\cite{benor2010ethnolinguistic}. This necessitates a more nuanced description for speakers and texts than a discrete dialect category.

Following prior work~\cite[e.g.,][]{vanhofwegen-wolfram-2010-ddm} 
we construct dialect density measures from feature detectors by counting the predicted number of features in each utterance, and dividing by the number of tokens. For the learning-based feature detectors (minimal pairs and corpus examples), we include partial counts from the detection probability; 
for the regular expression detectors, we simply count the number of matches and dividing by the number of tokens. In addition, we construct a DDM based on a document classifier: we train a classifier to distinguish Indian English from U.S. English, and then use its predictive probability as the DDM. These DDMs are then compared on two tasks: distinguishing Indian and U.S. English, and correlation with the density of expert-annotated features. The classifier is trained by fine-tuning BERT, using a prediction head on the \clstok token.

\subsection{Ranking documents by dialect density}
\label{sec:eval-density-ranking}
One application of dialect feature recognizers is to rank documents based on their dialect density, e.g. to identify challenging cases for evaluating downstream NLP systems, or for dialectology research. We correlate the dialect density against the density of expert-annotated features from \newcite{lange2012syntax}, both measured at the transcript-level, and report the Spearman rank-correlation $\rho$. 

As shown in \autoref{tab:ddm-performance}, the document classifier performs poorly: learning to distinguish Indian and U.S. English offers no information on the density of Indian dialect features, suggesting that the model is attending to other information, such as topics or entities. The feature-based model trained on labeled examples performs best, which is unsurprising because it is trained on the same type of features that it is now asked to predict. Performance is weaker when the model is trained from minimal pairs. Minimal pair training is particularly helpful on rare features, but offers far fewer examples on the high-frequency features, which in turn dominate the DDM scores on test data.
Regular expressions perform well on this task, because we happen to have regular expressions for the high-frequency features, and because the precision issues are less problematic in aggregate when the DDM is not applied to non-dialectal transcripts.

\subsection{Dialect Classification}
\label{sec:eval-density-classification}
Another application of dialect feature recognizers is to classify documents or passages by dialect~\cite{dunn2018finding}. This can help to test the performance of downstream models across dialects, assessing dialect transfer loss~\cite[e.g.,][]{blodgett2016demographic}, as well as identifying data of interest for manual dialectological research. We formulate a classification problem using the ICE-India and the Santa Barbara Corpus (ICE-USA). Each corpus is divided into equal-size training and test sets. The training corpus was also used for hyperparameter selection for the dialect feature recognition models, as described in \autoref{sec:classification-architectures}.

The dialect classifier was constructed by building on the components from \autoref{sec:eval-density-ranking}. For the test set, we measure the $D'$ (``D-prime'') statistic~\cite{macmillan1991detection}, 
\begin{equation}
D' = \frac{\mu_{\text{IN}} - \mu_{\text{US}}}{\sqrt{\frac{1}{2} (\sigma^2_{\text{IN}} + \sigma^2_{\text{US}})}}.
\end{equation} 
This statistic, which can be interpreted similarly to a $Z$-score, quantifies the extent to which a metric distinguishes between the two populations. 
We also report classification accuracy; lacking a clear way to set a threshold, for each classifier we balance the number of false positives and false negatives.

As shown in \autoref{tab:ddm-performance}, both the document classifier and the corpus-based feature detection model (trained on labeled examples) achieve high accuracy at discriminating U.S. and Indian English. The $D'$ discriminability score is higher for the document classifier, which is trained on a cross-entropy objective that encourages making confident predictions. Regular expressions suffer from low precision because they respond to surface cues that may be present in U.S. English, even when the dialect feature is not present (e.g., the word ``only'', the phrase ``is there'').

\begin{table}

    \centering
        \resizebox{\linewidth}{!}{%
    \begin{tabular}{lccc}
    \toprule
    & {\bf Ranking} & \multicolumn{2}{l}{\bf Classification}\\
    \cmidrule(lr){2-2}
     \cmidrule(lr){3-4}
    {\bf Dialect density measure} & $\rho$ & $D'$ & acc.  \\
    \midrule
        Document classifier & $-0.17$ & $14.48$ & $1$ \\
    Multihead, corpus examples &  $0.83$ &  $2.30$ & $0.95$ \\
    Multihead, minimal pairs & $0.70$ &  $1.85$ & $0.85$ \\
        Regular expressions &  $0.71$ &  $1.61$ & $0.80$ \\
         \bottomrule
    \end{tabular}%
    }
    \caption{Performance of dialect density measures at the tasks of ranking Indian English transcripts by dialect density (quantified by Spearman $\rho$) and distinguishing Indian and U.S. English transcripts (quantified by accuracy and $D'$ discriminability).
   }
    \label{tab:ddm-performance}
\end{table}

\section{Related Work}
\paragraph{Dialect classification.} Prior work on dialect in natural language processing has focused on distinguishing between dialects (and closely-related languages). For example, the VarDial 2014 shared task required systems to distinguish between nation-level language varieties, such as British versus U.S. English, as well as closely-related language pairs such as Indonesian versus Malay~\cite{zampieri-etal-2014-report}; later evaluation campaigns expanded this set to other varieties~\cite{zampieri-etal-2017-findings}. In general, participants in these shared tasks have taken a text classification approach; neural architectures have appeared in the more recent editions of these shared tasks, but with a few exceptions~\cite[e.g.,][]{bernier-colborne-etal-2019-improving}, they have not outperformed classical techniques such as support vector machines. Our work differs by focusing on a specific set of known dialect features, rather than document-level classification between dialects, which aligns with the linguistic view of dialects as bundles of correlated features~\cite{nerbonne2009data} and tracks variable realization of features within dialect usage.

\paragraph{Discovering and detecting dialect features.} 
Machine learning feature selection techniques have been employed to discover dialect features from corpora. For example, \newcite{dunn2018finding,dunn-2019-modeling} induces a set of \emph{constructions} (short sequences of words, parts-of-speech, or constituents) from a ``neutral'' corpus, and then identifies constructions with distinctive distributions over the geographical subcorpora of the International Corpus of English (ICE). In social media, features of African American Vernacular English (AAVE) can be identified by correlating linguistic frequencies with the aggregate demographic statistics of the geographical areas from which geotagged social media was posted~\cite{eisenstein-etal-2011-discovering,stewart-2014-now,blodgett2016demographic}. In contrast, we are interested in detecting predefined dialect features from well-validated resources such as dialect atlases. 

Along these lines, \newcite{jorgensen2015challenges} and \newcite{jones2015toward} designed lexical patterns to identify non-standard spellings that match known phonological variables from AAVE (e.g., \say{sholl} `sure'), demonstrating the presence of these variables in social media posts from regions with high proportions of African Americans.
\newcite{blodgett2016demographic} use the same geography-based approach to test for phonological spellings and constructions corresponding to syntactic variables such as habitual \say{be}; \newcite{hovy2015user} show that a syntactic feature of Jutland Danish can be linked to the geographical origin of product reviews. These approaches have focused mainly on features that could be recognized directly from surface forms, or in some cases, from part-of-speech (POS) sequences. In contrast, we show that it is possible to learn to recognize features from examples, enabling the recognition of features for which it is difficult or impossible to craft surface or POS patterns.



\paragraph{Minimal pairs in NLP.}  A distinguishing aspect of our approach is the use of minimal pairs rather than conventional labeled data. 
Minimal pairs are well known in natural language processing from the Winograd Schema~\cite{levesque2012winograd}, which is traditionally used for evaluation, but \newcite{kocijan2019surprisingly} show that fine-tuning on a related dataset of minimal pairs can improve performance on the Winograd Schema itself. A similar idea arises in counterfactually-augmented data~\cite{kaushik2019learning} and contrast sets~\cite{gardner2020evaluating}, in which annotators are asked to identify the minimal change to an example that is sufficient to alter its label. 
However, those approaches use counterfactual examples to \emph{augment} an existing training set, while we propose minimal pairs as a replacement for large-scale labeled data.
Minimal pairs have also been used to design controlled experiments and probe neural models' ability to capture various linguistic phenomena~\cite{gulordava2018colorless,ettinger2018assessing,futrell2019neural,gardner2020evaluating,schuster2020harnessing}. Finally, \newcite{liang2020alice} use contrastive explanations as part of an active learning framework to improve data efficiency. Our work shares the objective of \newcite{liang2020alice} to improve data efficiency, but is methodologically closer to probing work that uses minimal pairs to represent specific linguistic features.
\section{Conclusion}
We introduce the task of dialect feature detection and demonstrate that it is possible to construct dialect feature recognizers using only a small number of minimal pairs --- in most cases, just five positive and negative examples per feature. This makes it possible to apply computational analysis to the many dialects for which labeled data does not exist. 
Future work will extend this approach to multiple dialects, focusing on cases in which features are shared across two or more dialects. This lays the groundwork for the creation of dialect-based ``checklists''~\cite{ribeiro-etal-2020-beyond} to assess the performance of NLP systems across the diverse range of linguistic phenomena that may occur in any given language.

\section{Ethical Considerations} 

Our objective in building dialect feature recognizers is to aid developers and researchers to effectively benchmark NLP model performance across and within different dialects, and to assist social scientists and dialectologists studying dialect use. The capability to detect dialectal features may enable developers to test for and mitigate any unintentional and undesirable biases in their models towards or against individuals speaking particular dialects. This is especially important because dialect density has been documented to correlate with lower socioeconomic status \cite{sahgal-1988-inde}. However, this technology is not without its risks. As some dialects correlate with ethnicities or countries of origin, there is a potential dual use risk of the technology being used to profile individuals. Dialect features could also be used as predictors in downstream tasks; as with other proxies of demographic information, this could give the appearance of improving accuracy while introducing spurious correlations and imposing disparate impacts on disadvantaged groups. Hence we recommend that developers of this technology consider downstream use cases, including malicious use and misuse, when assessing the social impact of deploying and sharing this technology. 

The focus on predefined dialect features can introduce a potential source of bias if the feature set is oriented towards the speech of specific subcommunities within a dialect. However, analogous issues can arise in fully data-driven approaches, in which training corpora may also be biased towards subcommunities of speakers or writers. The feature-based approach has the advantage of making any such bias easier to identify and correct.

\paragraph{Acknowledgments.}
Thanks to Claudia Lange for sharing her annotations, and for discussion of this research. Thanks to Axel Bohmann for sharing information about his work on recognizing dialect features with regular expressions. Valuable feedback on this research was provided by Jason Baldridge, Dan Jurafsky, Slav Petrov, Jason Riesa, Kristina Toutanova, and especially Vera Axelrod. Thanks also to the anonymous reviewers. Devyani Sharma is supported in part by a Google Faculty Research Award. 

\bibliography{main}
\bibliographystyle{acl_natbib}

\appendix

\clearpage

\section{Regular Expressions}
\label{sec:regular_expressions}

\autoref{tab:regexes} shows the regular expressions that we used for the five features, where such patterns were available.

\begin{table*}
    \centering
    \resizebox{0.7\linewidth}{!}{%
\begin{tabular}{@{}ll@{}}
\toprule
\textbf{Feature}                           & \textbf{Regular expression}                                                                                                                                  \\ \midrule
\itself                             & \texttt{\textbackslash{}bitself\textbackslash{}b}                                                                                                                     \\
\only                               & \texttt{\textbackslash{}bonly\textbackslash{}b}                                                                                                                       \\
\existential & \texttt{\textbackslash{}bis there\textbackslash{}b|\textbackslash{}bare there\textbackslash{}b}                                                                       \\
\invtag     & \texttt{\textbackslash{}bisn't it\textbackslash{}b|\textbackslash{}bis it\textbackslash{}b|\textbackslash{}bno\textbackslash{}b|\textbackslash{}bna\textbackslash{}b} \\
\andall                          & \texttt{\textbackslash{}band all\textbackslash{}b}\\ \bottomrule
\end{tabular}%
}
    \caption{Regular expressions we used, for the features that such patterns were available.}
    \label{tab:regexes}
\end{table*}

\clearpage

\section{Sample Outputs}
\label{sec:appendix_outputs}

The examples below represent a random sample of the multihead models' outputs for Lange's features, comparing the one that is trained on corpus examples (\textsc{corpus}) to the one that is trained on minimal pairs (\textsc{minpair}). We show true positives (\textsc{tp}), false positives (\textsc{fp}) and false negatives (\textsc{fn}). We randomly sample three examples for each output type (\textsc{tp}, \textsc{fp}, \textsc{fn}) and model (\textsc{both}, \textsc{corpus} only, \textsc{minpair} only).

Our manual inspection shows a few errors in the human annotation by Lange and that certain false positives should be true positives, especially for \feature{\only}. We highlight such examples in \colorbox{lime}{green}. Among the rest of the false positives and false negatives, a large proportion of errors can be explained by contextual information that is not available to the models. For example, without context it is ambiguous whether ``we possess only'' is an example of \feature{\only}. Inspection of context shows that it is a truncated utterance, representing a standard use of \emph{only}, hence it is correctly characterized as a false positive. Another source of confusion to the model is missing punctuation. For example ``Both girls I have never left them alone till now'' could be construed as \feature{\ofronting} with \feature{\resobj}. However, in the original context, the example consists of multiple sentences: ``Two kids. Both girls. I have never left them alone till now.'' We removed punctuation from examples, since in many cases automatic ASR models do not produce punctuation either. However, this example demonstrates that punctuation can provide valuable information about clause and phrase boundaries, and should be included if possible.

\subsection{Focus \emph{itself}}
\begin{itemize}[label={},leftmargin=*,itemsep=0pt]
\item{[\textsc{tp:both}]}	We are feeling tired now itself
\item{[\textsc{tp:both}]}	Coach means they should be coached from when they are in nursery UKG itself
\item{[\textsc{tp:both}]}	I'm in final year but like they have started from first year itself
\item{[\textsc{tp:corpus}]}	And she got a chance of operating also during her internship itself nice and because that Cama hospital is for ladies only so she has lot of experience
\item{[\textsc{tp:minpair}]}	But even if they women is are working as much as a man she is earning the same monthly saving as a man itself
\item{[\textsc{tp:minpair}]}	You go around say one O'clock and then go for a movie and come back in the evening itself you see you
\item{[\textsc{fp:minpair}]}	And primarily you know the the issue orders were issued on fifth that is on the election day itself
\item{[\textsc{fp:minpair}]}	\colorbox{lime}{That is to we take on the coughs} \colorbox{lime}{our human blood itself}
\item{[\textsc{fp:minpair}]}	Now since you are doing the PGCT now after going back is it possible for you to use simple English in the classroom itself
\item{[\textsc{fn:both}]}	All the sums were there in the text book itself but still they have not done properly in the exam
\item{[\textsc{fn:both}]}	And thinking about dissection hall itself they really get scared and that also in the midnight
\item{[\textsc{fn:both}]}	Means what do you think that the basic itself is not good or now they are getting interest in maths
\item{[\textsc{fn:corpus}]}	But even if they women is are working as much as a man she is earning the same monthly saving as a man itself
\item{[\textsc{fn:corpus}]}	You go around say one O'clock and then go for a movie and come back in the evening itself you see you
\item{[\textsc{fn:minpair}]}	And she got a chance of operating also during her internship itself nice and because that Cama hospital is for ladies only so she has lot of experience
\end{itemize}
\begin{itemize}[label={},leftmargin=*,itemsep=0pt]
\subsection{Focus \emph{only}}
\item{[\textsc{tp:both}]}	All the types only
\item{[\textsc{tp:both}]}	Hey you sur be like that only
\item{[\textsc{tp:both}]}	suddenly it will be become perfect only
\item{[\textsc{tp:corpus}]}	That is I like dressing up I told you at the beginning only
\item{[\textsc{tp:corpus}]}	Because today only he had come and I've got up today at nine thirty
\item{[\textsc{tp:corpus}]}	Actually from childhood only I was brought up in the same atmosphere like if Papa still has shifted to another place I would have got the feeling of not having comfortable in a particular language but on the whole I think it doesn't matter exactly how we go about chosing or selecting a language
\item{[\textsc{tp:minpair}]}	it was bit it was difficult only
\item{[\textsc{tp:minpair}]}	I'm one minute I've got it in front of me only
\item{[\textsc{tp:minpair}]}	He is in our college only
\item{[\textsc{fp:both}]}	Because we are supposed to perform well there only then
\item{[\textsc{fp:both}]}	\colorbox{lime}{Ho Ho Hollywood Hollywood after} \colorbox{lime}{Hollywood it seems India only}
\item{[\textsc{fp:both}]}	\colorbox{lime}{No he'll be there in the campus only}
\item{[\textsc{fp:corpus}]}	\colorbox{lime}{Oh God there only it's happening} \colorbox{lime}{so and forget about}
\item{[\textsc{fp:corpus}]}	\colorbox{lime}{The thing is that it is rural area} \colorbox{lime}{only but the people are from all over india} \colorbox{lime}{they are staying here}
\item{[\textsc{fp:corpus}]}	Not much work these days because first week and last week only we've quiet good business
\item{[\textsc{fp:minpair}]}	Only in India there is manual work
\item{[\textsc{fp:minpair}]}	Film hits only
\item{[\textsc{fp:minpair}]}	\colorbox{lime}{So Bharati Vidya Bhavan people} \colorbox{lime}{have such type of persons only}
\item{[\textsc{fn:both}]}	If they be in always that this is there are not improve no improvement only
\item{[\textsc{fn:both}]}	When we were living when I was living in Kashmir no I was brought up there only and everything is
\item{[\textsc{fn:both}]}	This is the first phase then in the second phase we have some clinical subjects in which we come in direct contact with the patients but it's on two basis like when we see the patients at the same time we study about the pathology only the pathology and then we learn about some of the drugs which are to be which are used for their treatment
\item{[\textsc{fn:corpus}]}	No you must put apply science only
\item{[\textsc{fn:corpus}]}	Actually they are good only
\item{[\textsc{fn:corpus}]}	it was bit it was difficult only
\item{[\textsc{fn:minpair}]}	My both the parents are farmers only
\item{[\textsc{fn:minpair}]}	Because today only he had come and I've got up today at nine thirty
\item{[\textsc{fn:minpair}]}	That is I like dressing up I told you at the beginning only
\end{itemize}
\begin{itemize}[label={},leftmargin=*,itemsep=0pt]
\subsection{Invariant Tag (\emph{isn't it}, \emph{no}, \emph{na})}
\item{[\textsc{tp:both}]}	Very difficult once the school starts na very difficult
\item{[\textsc{tp:both}]}	I am okay rainy season no
\item{[\textsc{tp:both}]}	Oh yours your head is not reeling any more no ?
\item{[\textsc{tp:corpus}]}	Kind of but it would be better than an indoor game no
\item{[\textsc{tp:corpus}]}	We'll ask that person no that Sagar you can tell
\item{[\textsc{tp:corpus}]}	Nothing at all that's why you got scratching on that day I know that no that's why I asked
\item{[\textsc{tp:minpair}]}	I'm not fair no
\item{[\textsc{tp:minpair}]}	Husband no I'll do I'll prepare it
\item{[\textsc{tp:minpair}]}	He could have agreed no what is that
\item{[\textsc{fp:both}]}	TELCO deta hai to kuch problem nahi na
\item{[\textsc{fp:both}]}	\colorbox{lime}{I think once you have got in you no}
\item{[\textsc{fp:both}]}	\colorbox{lime}{I didn't go anywhere no}
\item{[\textsc{fp:corpus}]}	\colorbox{lime}{Or two hundred rupees that no}
\item{[\textsc{fp:corpus}]}	Know when we go back no I think we'll get a rosy welcome home welcome there
\item{[\textsc{fp:corpus}]}	I like straight and perspiration then only I feel at home otherwise no
\item{[\textsc{fp:minpair}]}	No got it repaired
\item{[\textsc{fp:minpair}]}	No no he is here
\item{[\textsc{fp:minpair}]}	Okay no but
\item{[\textsc{fn:both}]}	I just go out for tea isn't
\item{[\textsc{fn:both}]}	Hey you you like serious movies is it you like serious movies
\item{[\textsc{fn:both}]}	See no the scene exactly happened you know the other day what happen I was reading baba
\item{[\textsc{fn:corpus}]}	I'm not fair no
\item{[\textsc{fn:corpus}]}	I think no
\item{[\textsc{fn:corpus}]}	Tell me no why you can't tell
\item{[\textsc{fn:minpair}]}	Yeah then it's first time first time it was new to me no
\item{[\textsc{fn:minpair}]}	That is the main thing na here that would again the main thing that they don't take at all interest in the their children at all
\item{[\textsc{fn:minpair}]}	So culture nahi hai there is I don't follow culture religion nothing na
\end{itemize}
\begin{itemize}[label={},leftmargin=*,itemsep=0pt]
\subsection{Lack of Copula}
\item{[\textsc{fp:corpus}]}	Which October first I think
\item{[\textsc{fp:corpus}]}	June nineteen eighty-six
\item{[\textsc{fp:minpair}]}	Construction all before
\item{[\textsc{fp:minpair}]}	Not in the class
\item{[\textsc{fp:minpair}]}	The tendency to
\item{[\textsc{fn:both}]}	you've she said his grandfather still working
\item{[\textsc{fn:both}]}	Everybody so worried about the exams and studies
\item{[\textsc{fn:both}]}	Again classes bit too long I feel five O'clock is tiring
\end{itemize}
\begin{itemize}[label={},leftmargin=*,itemsep=0pt]
\subsection{Left Dislocation}
\item{[\textsc{tp:both}]}	This principal she is very particular about it
\item{[\textsc{tp:both}]}	Vilas and Ramesh they they make noise man
\item{[\textsc{tp:both}]}	That's why those Muslims they got very angry
\item{[\textsc{tp:corpus}]}	And med medium class they can't understand soon
\item{[\textsc{tp:corpus}]}	That will become difficult and common people they don't understand
\item{[\textsc{tp:corpus}]}	And now the Kukis they refused to pay any more
\item{[\textsc{tp:minpair}]}	It's because of this some other participant they complained about this and then they started they started this particular
\item{[\textsc{tp:minpair}]}	We've lot of fun in theatres you know we always take the back seat and all that for this guys distinct one we keep teasing them
\item{[\textsc{tp:minpair}]}	My post graduation degree I finished it in mid June nineteen eighty-six
\item{[\textsc{fp:both}]}	But whereas when they really come to know the people they like to help the people
\item{[\textsc{fp:both}]}	It's actually some of them like to see it really so huge and long and bigger snakes they are in all closed and all there it is nice to see it
\item{[\textsc{fp:both}]}	But generally the educated people I don't find much variation but in accent there may be a variation
\item{[\textsc{fp:corpus}]}	Everytime he keeps speaking you know they get irritated and say aram se
\item{[\textsc{fp:corpus}]}	What happened is they will change programme and the fifty guys they'll just keep quite
\item{[\textsc{fp:corpus}]}	Whereas Hyderabad the people are more conservative and like they don't like to go out even or at the first move they don't like to talk with people also
\item{[\textsc{fp:minpair}]}	And the songs now once we hear it afterwards when some other famous songs comes that we forget the last ones
\item{[\textsc{fp:minpair}]}	But when we approach since it seems they they put lot of conditions yes that you fed up with those people and
\item{[\textsc{fp:minpair}]}	so that's why we missed we that missed that holiday it being a Sunday
\item{[\textsc{fn:both}]}	Administration it is all done by Bharati Vidya Bhavan
\item{[\textsc{fn:both}]}	Oh our Joshi okay II got got him
\item{[\textsc{fn:both}]}	Yes yes it is true but our constitution makers
\item{[\textsc{fn:corpus}]}	and he has used the the place where the palace once palace might be there and that portion and the remaining part he built an antenna he has fixed it there at the top
\item{[\textsc{fn:corpus}]}	Not exactly but Calcutta sweets I think they do have a little flavour and that I haven't got anywhere in India
\item{[\textsc{fn:corpus}]}	Computer it was in the first semester
\item{[\textsc{fn:minpair}]}	And med medium class they can't understand soon
\item{[\textsc{fn:minpair}]}	Shireen she was excellent at that
\item{[\textsc{fn:minpair}]}	Yeah arti arti students they loiter about in the corridor
\end{itemize}
\begin{itemize}[label={},leftmargin=*,itemsep=0pt]

\subsection{Non-initial Existential \emph{X is / are there}}
\item{[\textsc{tp:both}]}	Libraries are there
\item{[\textsc{tp:both}]}	only specimen like operated cases like supposing a is there
\item{[\textsc{tp:both}]}	Problems are there problems are there what
\item{[\textsc{tp:corpus}]}	to assist there some teachers are there and together we conduct the classes
\item{[\textsc{tp:corpus}]}	It's there but it's common no
\item{[\textsc{tp:corpus}]}	Yeah I think Varlaxmi is there
\item{[\textsc{fp:both}]}	My husband is there mother is there
\item{[\textsc{fp:corpus}]}	Come no Shaukat is here Natalie is here even if Savita is not there they two are there na
\item{[\textsc{fp:corpus}]}	Actually there the thing is that you know for example
\item{[\textsc{fp:corpus}]}	Any thing is there produced materials which do not require much resource personnel
\item{[\textsc{fp:minpair}]}	Ph D degree is awarded there
\item{[\textsc{fn:both}]}	Yeah the royalties too there they're there and we've the king
\item{[\textsc{fn:both}]}	Okay somebody else's some somebody else is there
\item{[\textsc{fn:both}]}	In that you know everything is about nature I'll tell you yeah it's very lovely means very nice lovely what but and small children were there in that
\item{[\textsc{fn:minpair}]}	American and all other capitalist nations were also there
\item{[\textsc{fn:minpair}]}	Nice movie yaar that song is there no hai apna dil to awara
\item{[\textsc{fn:minpair}]}	It's not there
\end{itemize}
\begin{itemize}[label={},leftmargin=*,itemsep=0pt]

\subsection{Object Fronting}
\item{[\textsc{tp:both}]}	Just typing work I have to do
\item{[\textsc{tp:corpus}]}	writing skills there are so many you can teach them
\item{[\textsc{tp:corpus}]}	Each other and so many things we have learnt
\item{[\textsc{tp:corpus}]}	My birthday party you arrange
\item{[\textsc{fp:corpus}]}	Formalities I will come
\item{[\textsc{fp:corpus}]}	Mar Marxism you were
\item{[\textsc{fp:minpair}]}	Other wise we have to
\item{[\textsc{fn:both}]}	That also I'm not having just I jump jumped jumped I came studies also
\item{[\textsc{fn:both}]}	Yes Hawa Mahal we heard
\item{[\textsc{fn:both}]}	About ten to twenty books I'll read that's all
\item{[\textsc{fn:minpair}]}	Small baby very nice it was
\item{[\textsc{fn:minpair}]}	But more keen she is
\item{[\textsc{fn:minpair}]}	And camera handling actually outdoor landscaping that landscape shot I have taken and actually the close ups and some parts of your architectural shots of that building Ganesh took my husband took and close ups of the faces my husband and Ganesh took
\end{itemize}
\begin{itemize}[label={},leftmargin=*,itemsep=0pt]

\subsection{Resumptive Object Pronoun}
\item{[\textsc{tp:minpair}]}	and he has used the the place where the palace once palace might be there and that portion and the remaining part he built an antenna he has fixed it there at the top
\item{[\textsc{tp:minpair}]}	Yeah also pickles we eat it with this jaggery and lot of butter
\item{[\textsc{tp:minpair}]}	My post graduation degree I finished it in mid June nineteen eighty-six
\item{[\textsc{fp:minpair}]}	Having humurous something special I would love it to join it
\item{[\textsc{fp:minpair}]}	I see a number of people I like them very much
\item{[\textsc{fp:minpair}]}	Old and ancient things in carving we get it so beautifully
\item{[\textsc{fn:both}]}	Oh our Joshi okay II got got him
\item{[\textsc{fn:both}]}	Normaly no we don't overdrawn on account but haan haan whatever is balance you know yeah help them give them suppose cheque books and all we are supposed to keep them yeah two fifty balance
\item{[\textsc{fn:both}]}	He is in a that's what he was telling me today see I want your draft like draft draft by January by the month of January by the end of January so that II might rectify it and then I will do it I will give it back to you by mid Febraury so that you can get it final draft by by the end of Febraury
\item{[\textsc{fn:corpus}]}	and he has used the the place where the palace once palace might be there and that portion and the remaining part he built an antenna he has fixed it there at the top
\item{[\textsc{fn:corpus}]}	Yeah also pickles we eat it with this jaggery and lot of butter
\item{[\textsc{fn:corpus}]}	My post graduation degree I finished it in mid June nineteen eighty-six
\end{itemize}
\begin{itemize}[label={},leftmargin=*,itemsep=0pt]
\subsection{Resumptive Subject Pronoun}
\item{[\textsc{tp:corpus}]}	Like those terrorists they wanted us to to accompany them in the revolt against India
\item{[\textsc{tp:corpus}]}	And one more thing another thing how I rectified myself because all almost all all of us all my brother and sisters we have read in English medium school
\item{[\textsc{tp:corpus}]}	Dr this Mr V he was totally changed actually because he was the concepts are clear not clear to us
\item{[\textsc{fp:corpus}]}	There are so many people they can they could shine like anything
\item{[\textsc{fp:corpus}]}	Kolhapur he had come to Guwahati
\item{[\textsc{fp:corpus}]}	I don't know what he whenever whenever I see those guys they they nicely speak to me
\item{[\textsc{fp:minpair}]}	His house he is going to college KK diploma electronics
\item{[\textsc{fn:both}]}	they I thought that another one Patil is there a horrible he is I thought that Patil
\item{[\textsc{fn:both}]}	Computer it it plays a great role because we are having computers in each field now-a-days
\item{[\textsc{fn:both}]}	You know that a woman she is a apprehensive about many things
\item{[\textsc{fn:minpair}]}	Like those terrorists they wanted us to to accompany them in the revolt against India
\item{[\textsc{fn:minpair}]}	Whereas in Hyderabad they still have the old cultures and so many things that even the parents they don't even let the girls talk with the guys
\item{[\textsc{fn:minpair}]}	And the students who come out with a degree MMSI understand that there is a report that has been received from different firms that the students of BITS Pilani specially MMS candidates they are prepared to soil their hands
\end{itemize}
\begin{itemize}[label={},leftmargin=*,itemsep=0pt]
\subsection{Topicalized Non-argument Constituent}
\item{[\textsc{tp:corpus}]}	for Diwali you went I know that
\item{[\textsc{tp:corpus}]}	So very long time we have not travelled together
\item{[\textsc{tp:corpus}]}	Pooja vacation also we used to conduct some classes practical classes
\item{[\textsc{tp:minpair}]}	In pooja day some important days we stay back
\item{[\textsc{fp:corpus}]}	In Jaipur then we have also we have a Birla
\item{[\textsc{fp:corpus}]}	Like that we
\item{[\textsc{fp:corpus}]}	Everytime we have some work to do
\item{[\textsc{fp:minpair}]}	Aa i i initial periods I did very difficult but I
\item{[\textsc{fn:both}]}	I mean here in Hyderabad the people are it's okay they are nice
\item{[\textsc{fn:both}]}	And that old ones again we put them we feel like hearing again
\item{[\textsc{fn:both}]}	But in drama we'll have to be very different
\item{[\textsc{fn:corpus}]}	In pooja day some important days we stay back
\item{[\textsc{fn:minpair}]}	for Diwali you went I know that
\item{[\textsc{fn:minpair}]}	Pooja vacation also we used to conduct some classes practical classes
\item{[\textsc{fn:minpair}]}	Sir from Monday onwards I too want to take leave sir for four days because total I have five C Ls so from
\end{itemize}

\clearpage

\section{Average Precision Results}
\label{app:avg-prec}

\begin{table}[h]
\pgfkeys{/pgf/number format/.cd,fixed,fixed zerofill,precision=3}
\setlength{\tabcolsep}{0.5ex}
\resizebox{\linewidth}{!}{%
\begin{filecontents*}{tables/lange_AvgPrecision.csv}
shortnames,finetune_annotated_data_lange,finetune_annotated_data_multilabel_lange,finetune_minimal_pairs_lange,finetune_minimal_pairs_multilabel_lange
\itself*,0.6676239466269156,0.6308265473645358,0.664909809095979,0.6125952470559028
\only*,0.582074013608971,0.4037800242474082,0.34374188261272876,0.4160740354536688
\textsc{invariant tag},0.8760124115535554,0.8707233581730609,0.44122096492499113,0.49455368016096124
\copula,0.02919396917989947,0.014744789288238627,0.011673738279115458,0.03554973441981664
\leftdis,0.4252705835817297,0.38270625842900524,0.14931468523731575,0.2317939772268748
\existential*,0.8865941467940683,0.9063982314144358,0.5559910066930461,0.5099476428834844
\ofronting,0.23757160986761966,0.20160583663329676,0.030856689843086642,0.08252485958865781
\textsc{res. object pronoun},0.052439349980433314,0.019905231213874452,0.04597687339195521,0.06127620841980086
\textsc{res. subject pronoun},0.4601539494356877,0.40859983341158673,0.0776205034138526,0.19848966383667369
\textsc{topicalized non-arg. const.},0.07968000357150941,0.07593288366247755,0.021156958014846794,0.044336782798063024
\textbf{Macro Average},0.4296613984200389,0.391522299383792,0.23424631115069175,0.2687141831843904
\end{filecontents*}
\pgfplotstabletypeset[
    header=true,
    col sep=comma,
    white space chars={_},
    column type={lcccc},
    columns/shortnames/.style={string type,column name=Dialect feature},
    columns/finetune annotated data multilabel lange/.style={column name=Multihead},
    columns/finetune annotated data lange/.style={column name=\damodel},
    columns/finetune minimal pairs multilabel lange/.style={column name=Multihead},
    columns/finetune minimal pairs lange/.style={column name=\damodel},
    every head row/.style={
    before row={\toprule {\bf Supervision:} & \multicolumn{2}{c}{\bf Corpus examples} & \multicolumn{2}{c}{\bf Minimal pairs}\\},
    after row=\midrule},
    every last row/.style={after row=\bottomrule},    ]{tables/lange_AvgPrecision.csv}
}
\caption{Average precision for the Lange features. Scores are in the range $[0, 1]$, with $1$ indicating perfect performance. Asterisks mark features that can be recognized with a regular expression.}
\end{table}

\begin{table}[h]
\pgfkeys{/pgf/number format/.cd,fixed,fixed zerofill,precision=3}
\setlength{\tabcolsep}{0.5ex}
\resizebox{\linewidth}{!}{%
\begin{filecontents*}{tables/ours_AvgPrecision.csv}
shortnames,finetune_minimal_pairs_multilabel_ours,finetune_minimal_pairs_ours
\article,0.3078846422658458,0.21024096757671823
\dodrop,0.05724879741959986,0.04433135805554686
\extraarticle,0.06474818058655818,0.11648356673601505
\itself*,0.853141578963512,1.0
\only*,0.27438427656138453,0.8587003968253969
\habitual,0.019563262576968155,0.008166534487983358
\textsc{invariant tag},0.4204019805920411,0.6143001138295257
\embedded,0.16156988678677672,0.1057943072328769
\agreement,0.11007017172903526,0.0836519519244058
\whquestions,0.10608917881495358,0.3092708972281065
\leftdis,0.3006841489704294,0.2878896222220199
\massnouns,0.034462782299954045,0.045064022703373605
\existential*,0.39744277171522935,0.5057351453445236
\ofronting,0.19269807562539043,0.1470057822830329
\prepdrop,0.1158300191130053,0.06378279235145226
\ppfronting,0.13442071124433802,0.09068088120584747
\stative,0.32912137602212066,0.26678954202908184
\andall,0.7781756338899195,0.7693970921702014
\textbf{Macro Average},0.25877430417650343,0.30707138745589496
\end{filecontents*}
\pgfplotstabletypeset[
    header=true,
    col sep=comma,
    white space chars={_},
    column type={lcc},
    columns={shortnames,finetune minimal pairs ours,finetune minimal pairs multilabel ours},
    columns/shortnames/.style={string type,column name={\bf Dialect feature}},
        columns/finetune minimal pairs ours/.style={column name={\bf \damodel}},
    columns/finetune minimal pairs multilabel ours/.style={column name={\bf Multihead}},
    every head row/.style={
    before row={\toprule}, after row=\midrule},
    every last row/.style={after row=\bottomrule},    ]{tables/ours_AvgPrecision.csv}
}
\caption{Average precision for the extended feature set. As described in the main text, corpus training examples are unavailable for these features.}
\end{table}

\clearpage
\onecolumn
\section{Minimal pairs}
\label{sec:appendix_minpairs}
\footnotesize
\begin{longtable}{cllr}
\multicolumn{1}{l}{\textbf{ID}} & \textbf{Feature}                                        & \textbf{Example}                                                                                                                                           & \multicolumn{1}{l}{\textbf{Label}} \\
\hline
\endfirsthead
\endhead
1                               & \article                                        & the person I like the most is from mechanical department                                                                                                   & 1                                  \\
1                               & \article                                        & person I like the most is from the mechanical department                                                                                                   & 1                                  \\
1                               & \article                                        & person I like most is from the mechanical department                                                                                                       & 1                                  \\
1                               & \article                                        & person I like most is from mechanical department                                                                                                           & 1                                  \\
1                               & \article                                        & the person I like the most is from the mechanical department                                                                                               & 0                                  \\
2                               & \article                                        & we can only see blue sky                                                                                                                                   & 1                                  \\
2                               & \article                                        & we can only see the blue sky                                                                                                                               & 0                                  \\
3                               & \article                                        & recipe is simple thing                                                                                                                                     & 1                                  \\
3                               & \article                                        & recipe is a simple thing                                                                                                                                   & 1                                  \\
3                               & \article                                        & a recipe is simple thing                                                                                                                                   & 1                                  \\
3                               & \article                                        & a recipe is a simple thing                                                                                                                                 & 0                                  \\
4                               & \article                                        & union person contacted his representative at the school                                                                                                    & 1                                  \\
4                               & \article                                        & the union person contacted his representative at the school                                                                                                & 0                                  \\
5                               & \article                                        & it was first day of term                                                                                                                                   & 1                                  \\
5                               & \article                                        & it was the first day of term                                                                                                                               & 0                                  \\
6                               & \dodrop                                   & we have two tailors who can make for us                                                                                                                    & 1                                  \\
6                               & \dodrop                                   & we have two tailors who can make clothes for us                                                                                                            & 0                                  \\
6                               & \dodrop                                   & we have two tailors who can make them for us                                                                                                               & 0                                  \\
7                               & \dodrop                                   & he didn't give me                                                                                                                                          & 1                                  \\
7                               & \dodrop                                   & he didn't give it to me                                                                                                                                    & 0                                  \\
8                               & \dodrop                                   & in our old age we can go and enjoy                                                                                                                         & 1                                  \\
8                               & \dodrop                                   & in our old age we can go and enjoy it                                                                                                                      & 0                                  \\
9                               & \dodrop                                   & she doesn't like                                                                                                                                           & 1                                  \\
9                               & \dodrop                                   & she doesn't like it                                                                                                                                        & 0                                  \\
10                              & \dodrop                                   & he likes here more                                                                                                                                         & 1                                  \\
10                              & \dodrop                                   & he likes it here more                                                                                                                                      & 0                                  \\
11                              & \itself                                          & \begin{tabular}[c]{@{}l@{}}So if you're not good at communication you may get filtered\\ at the first level itself\end{tabular}                            & 1                                  \\
11                              & \itself                                          & \begin{tabular}[c]{@{}l@{}}So if you're not good at communication you may get filtered\\ at even the first level\end{tabular}                              & 0                                  \\
12                              & \itself                                          & \begin{tabular}[c]{@{}l@{}}But I did have some difficulty getting to know people among\\ Indians itself\end{tabular}                                       & 1                                  \\
12                              & \itself                                          & \begin{tabular}[c]{@{}l@{}}But I did have some difficulty getting to know people among\\ Indians themselves\end{tabular}                                   & 0                                  \\
13                              & \itself                                          & I think you should start going to the gym from now itself.                                                                                                 & 1                                  \\
13                              & \itself                                          & I think you should start going to the gym from now.                                                                                                        & 0                                  \\
14                              & \itself                                          & I did one refresher course in the month of June itself.                                                                                                    & 1                                  \\
14                              & \itself                                          & I did one refresher course in the month of June.                                                                                                           & 0                                  \\
15                              & \itself                                          & He is doing Engineering in Delhi itself.                                                                                                                   & 1                                  \\
15                              & \itself                                          & He is doing Engineering in Delhi.                                                                                                                          & 0                                  \\
16                              & \only                                            & I'm working very nearby to my house only                                                                                                                   & 1                                  \\
16                              & \only                                            & I'm working very near my house                                                                                                                             & 0                                  \\
17                              & \only                                            & recently only in April there was a big fight                                                                                                               & 1                                  \\
17                              & \only                                            & as recently as April there was a big fight                                                                                                                 & 0                                  \\
18                              & \only                                            & I was there yesterday only                                                                                                                                 & 1                                  \\
18                              & \only                                            & I was there just yesterday                                                                                                                                 & 0                                  \\
19                              & \only                                            & She was brought up there and her college was there only                                                                                                    & 1                                  \\
19                              & \only                                            & She was brought up there and her college was there too                                                                                                     & 0                                  \\
20                              & \only                                            & You get on the train and buy the ticket there only                                                                                                         & 1                                  \\
20                              & \only                                            & You get on the train and buy the ticket there too                                                                                                          & 0                                  \\
21                              & \habitual                                    & anybody giving donation, we are giving receipt                                                                                                             & 1                                  \\
21                              & \habitual                                    & if anybody gives a donation, we give a receipt                                                                                                             & 0                                  \\
22                              & \habitual                                    & she is getting nightmares                                                                                                                                  & 1                                  \\
22                              & \habitual                                    & she gets nightmares                                                                                                                                        & 0                                  \\
23                              & \habitual                                    & they are getting H1B visas to come to the country                                                                                                          & 1                                  \\
23                              & \habitual                                    & they get H1B visas to come to the country                                                                                                                  & 0                                  \\
24                              & \habitual                                    & they are teasing the new children when they join                                                                                                           & 1                                  \\
24                              & \habitual                                    & they tease the new children when they join                                                                                                                 & 0                                  \\
25                              & \habitual                                    & everyone is getting that vaccination in childhood                                                                                                          & 1                                  \\
25                              & \habitual                                    & everyone gets that vaccination in childhood                                                                                                                & 0                                  \\
26                              & \invtag                  & the children are playing outside, isn't it?                                                                                                                & 1                                  \\
26                              & \invtag                  & the children are playing outside, no?                                                                                                                      & 1                                  \\
26                              & \invtag                  & the children are playing outside, na?                                                                                                                      & 1                                  \\
26                              & \invtag                  & the children are playing outside, aren't they?                                                                                                             & 0                                  \\
27                              & \invtag                  & I was very scared to, no?                                                                                                                                  & 1                                  \\
27                              & \invtag                  & I was very scared to, na?                                                                                                                                  & 1                                  \\
27                              & \invtag                  & I was very scared to, wasn't I?                                                                                                                            & 0                                  \\
28                              & \invtag                  & the store is around the corner, no, by the post office                                                                                                     & 1                                  \\
28                              & \invtag                  & the store is around the corner, na, by the post office                                                                                                     & 1                                  \\
28                              & \invtag                  & the store is around the corner by the post office                                                                                                          & 0                                  \\
29                              & \invtag                  & It’s come from me, no?                                                                                                                                     & 1                                  \\
29                              & \invtag                  & It’s come from me, na?                                                                                                                                     & 1                                  \\
29                              & \invtag                  & It’s come from me, hasn't it?                                                                                                                              & 0                                  \\
30                              & \invtag                  & he liked it, no, even though you said he wouldn't                                                                                                          & 1                                  \\
30                              & \invtag                  & he liked it, na, even though you said he wouldn't                                                                                                          & 1                                  \\
30                              & \invtag                  & he liked it, right, even though you said he wouldn't                                                                                                       & 0                                  \\
30                              & \invtag                  & he liked it, didn't he, even though you said he wouldn't                                                                                                   & 0                                  \\
31                              & \embedded                            & you cannot ask them why are they not coming for clinic visits                                                                                              & 1                                  \\
31                              & \embedded                            & you cannot ask them why they are not coming for clinic visits                                                                                              & 0                                  \\
32                              & \embedded                            & I don't know now what are they doing                                                                                                                       & 1                                  \\
32                              & \embedded                            & I don't know now what they are doing                                                                                                                       & 0                                  \\
33                              & \embedded                            & he was wondering why did the police stop him                                                                                                               & 1                                  \\
33                              & \embedded                            & he was wondering why the police stopped him                                                                                                                & 0                                  \\
34                              & \embedded                            & we want to know how can we make your favorite dish                                                                                                         & 1                                  \\
34                              & \embedded                            & we want to know how we can make your favorite dish                                                                                                         & 0                                  \\
35                              & \embedded                            & the school principal called me to ask when are you going back                                                                                              & 1                                  \\
35                              & \embedded                            & the school principal called me to ask when you are going back                                                                                              & 0                                  \\
36                              & \agreement                                       & he do a lot of things                                                                                                                                      & 1                                  \\
36                              & \agreement                                       & he does a lot of things                                                                                                                                    & 0                                  \\
37                              & \agreement                                       & my bother said that one of his favorite place is the beach nearby                                                                                          & 1                                  \\
37                              & \agreement                                       & my bother said that one of his favorite places is the beach nearby                                                                                         & 0                                  \\
38                              & \agreement                                       & only his shoes is visible                                                                                                                                  & 1                                  \\
38                              & \agreement                                       & only his shoes are visible                                                                                                                                 & 0                                  \\
39                              & \agreement                                       & \begin{tabular}[c]{@{}l@{}}ten years ago you didn't operated a machine that could lift\\ a house all by itself\end{tabular}                                & 1                                  \\
39                              & \agreement                                       & \begin{tabular}[c]{@{}l@{}}ten years ago you didn't operate a machine that could lift\\ a house all by itself\end{tabular}                                 & 0                                  \\
40                              & \agreement                                       & he talk to them                                                                                                                                            & 1                                  \\
40                              & \agreement                                       & he talks to them                                                                                                                                           & 0                                  \\
41                              & \textsc{lack of inv. in wh-questions}                       & where you will get anything?                                                                                                                               & 1                                  \\
41                              & \textsc{lack of inv. in wh-questions}                       & where will you get anything?                                                                                                                               & 0                                  \\
42                              & \textsc{lack of inv. in wh-questions}                       & what you are doing?                                                                                                                                        & 1                                  \\
42                              & \textsc{lack of inv. in wh-questions}                       & what are you doing?                                                                                                                                        & 0                                  \\
43                              & \textsc{lack of inv. in wh-questions}                       & why you are telling this to everybody?                                                                                                                     & 1                                  \\
43                              & \textsc{lack of inv. in wh-questions}                       & why are you telling this to everybody?                                                                                                                     & 0                                  \\
44                              & \textsc{lack of inv. in wh-questions}                       & why you are driving like a lorry?                                                                                                                          & 1                                  \\
44                              & \textsc{lack of inv. in wh-questions}                       & why are you driving like a lorry?                                                                                                                          & 0                                  \\
45                              & \textsc{lack of inv. in wh-questions}                       & how your mother is feeling?                                                                                                                                & 1                                  \\
45                              & \textsc{lack of inv. in wh-questions}                       & how is your mother feeling?                                                                                                                                & 0                                  \\
46                              & \leftdis                                        & my father, he works for a mining company                                                                                                                   & 1                                  \\
46                              & \leftdis                                        & my father works for a mining company                                                                                                                       & 0                                  \\
47                              & \leftdis                                        & nowadays all the children they are mature from a very early age                                                                                            & 1                                  \\
47                              & \leftdis                                        & nowadays all the children are mature from a very early age                                                                                                 & 0                                  \\
48                              & \leftdis                                        & the camera, the dog is facing towards it                                                                                                                   & 1                                  \\
48                              & \leftdis                                        & the dog is facing towards the camera                                                                                                                       & 0                                  \\
49                              & \leftdis                                        & and all the company people, they are my clients                                                                                                            & 1                                  \\
49                              & \leftdis                                        & and all the company people are my clients                                                                                                                  & 0                                  \\
50                              & \leftdis                                        & those who come here definitely they should learn English                                                                                                   & 1                                  \\
50                              & \leftdis                                        & those who come here should definitely learn English                                                                                                        & 0                                  \\
51                              & \massnouns                               & this is a menial work                                                                                                                                      & 1                                  \\
51                              & \massnouns                               & this is menial work                                                                                                                                        & 0                                  \\
52                              & \massnouns                               & open a shop wherever there is a foot traffic                                                                                                               & 1                                  \\
52                              & \massnouns                               & open a shop wherever there is foot traffic                                                                                                                 & 0                                  \\
53                              & \massnouns                               & all the musics are very good                                                                                                                               & 1                                  \\
53                              & \massnouns                               & all the music is very good                                                                                                                                 & 0                                  \\
54                              & \massnouns                               & some informations are available free                                                                                                                       & 1                                  \\
54                              & \massnouns                               & some information is available free                                                                                                                         & 0                                  \\
55                              & \massnouns                               & they use proper grammars there                                                                                                                             & 1                                  \\
55                              & \massnouns                               & they use proper grammar there                                                                                                                              & 0                                  \\
56                              & \existential              & some flower part is there                                                                                                                                  & 1                                  \\
56                              & \existential              & there is some flower part                                                                                                                                  & 0                                  \\
57                              & \existential              & corruption is there obviously                                                                                                                              & 1                                  \\
57                              & \existential              & there is corruption obviously                                                                                                                              & 0                                  \\
58                              & \existential              & because in India individuality is not there                                                                                                                & 1                                  \\
58                              & \existential              & because there is no individuality in India                                                                                                                 & 0                                  \\
59                              & \existential              & five balls are there                                                                                                                                       & 1                                  \\
59                              & \existential              & there are five balls                                                                                                                                       & 0                                  \\
60                              & \existential              & every year inflation is there                                                                                                                              & 1                                  \\
60                              & \existential              & every year there is inflation                                                                                                                              & 0                                  \\
61                              & \ofronting                                         & not so much adjustment i have to make                                                                                                                      & 1                                  \\
61                              & \ofronting                                         & i don't have to make so much adjustment                                                                                                                    & 0                                  \\
61                              & \ofronting                                         & i have to make not so much adjustment                                                                                                                      & 0                                  \\
62                              & \ofronting                                         & minimum one month you have to wait                                                                                                                         & 1                                  \\
62                              & \ofronting                                         & you have to wait a minimum of one month                                                                                                                    & 0                                  \\
63                              & \ofronting                                         & Hindi Gujarati and Marathi you can use in Bombay                                                                                                           & 1                                  \\
63                              & \ofronting                                         & you can use Hindi Gujarati and Marathi in Bombay                                                                                                           & 0                                  \\
64                              & \ofronting                                         & in fifteen years lot of changes we have seen                                                                                                               & 1                                  \\
64                              & \ofronting                                         & in fifteen years we have seen a lot of changes                                                                                                             & 0                                  \\
65                              & \ofronting                                         & tomorrow this cake you have to try                                                                                                                         & 1                                  \\
65                              & \ofronting                                         & tomorrow you have to try this cake                                                                                                                         & 0                                  \\
66                              & \prepdrop                                        & I can see some green colour leaves the left side                                                                                                           & 1                                  \\
66                              & \prepdrop                                        & I can see some green colour leaves on the left side                                                                                                        & 0                                  \\
67                              & \prepdrop                                        & I went one year there.                                                                                                                                     & 1                                  \\
67                              & \prepdrop                                        & I went there for one year.                                                                                                                                 & 0                                  \\
68                              & \prepdrop                                        & We don’t feel that we should go any other country.                                                                                                         & 1                                  \\
68                              & \prepdrop                                        & We don’t feel that we should go to any other country.                                                                                                      & 0                                  \\
69                              & \prepdrop                                        & Those days it was considered a good job.                                                                                                                   & 1                                  \\
69                              & \prepdrop                                        & In those days it was considered a good job.                                                                                                                & 0                                  \\
70                              & \prepdrop                                        & So that time they said okay go and work for a few months.                                                                                                  & 1                                  \\
70                              & \prepdrop                                        & So at that time they said okay go and work for a few months.                                                                                               & 0                                  \\
71                              & \ppfronting & first of all, right side we can see a plate                                                                                                                & 1                                  \\
71                              & \ppfronting & first of all, we can see a plate the right side                                                                                                            & 0                                  \\
71                              & \ppfronting & first of all, we can see a plate on the right side                                                                                                         & 0                                  \\
71                              & \ppfronting & first of all, on the right side we can see a plate                                                                                                         & 0                                  \\
71                              & \article                                        & first of all, right side we can see a plate                                                                                                                & 1                                  \\
71                              & \prepdrop                                        & first of all, right side we can see a plate                                                                                                                & 1                                  \\
71                              & \prepdrop                                        & first of all, we can see a plate the right side                                                                                                            & 1                                  \\
72                              & \ppfronting & Tirupati temple I stayed one or two days                                                                                                                   & 1                                  \\
72                              & \ppfronting & I stayed one or two days at the Tirupati temple                                                                                                            & 0                                  \\
72                              & \ppfronting & at the Tirupati temple I stayed one or two days                                                                                                            & 0                                  \\
72                              & \article                                        & Tirupati temple I stayed one or two days                                                                                                                   & 1                                  \\
72                              & \prepdrop                                        & Tirupati temple I stayed one or two days                                                                                                                   & 1                                  \\
73                              & \ppfronting & two years I stayed alone                                                                                                                                   & 1                                  \\
73                              & \ppfronting & for two years I stayed alone                                                                                                                               & 0                                  \\
73                              & \prepdrop                                        & two years I stayed alone                                                                                                                                   & 1                                  \\
74                              & \ppfronting & you can say anything but tenth I'm leaving                                                                                                                 & 1                                  \\
74                              & \ppfronting & you can say anything but on the tenth I'm leaving                                                                                                          & 0                                  \\
74                              & \article                                        & you can say anything but tenth I'm leaving                                                                                                                 & 1                                  \\
74                              & \prepdrop                                        & you can say anything but tenth I'm leaving                                                                                                                 & 1                                  \\
75                              & \ppfronting & actually, this part I have not been                                                                                                                        & 1                                  \\
75                              & \ppfronting & actually, I have not been to this part                                                                                                                     & 0                                  \\
75                              & \prepdrop                                        & actually, this part I have not been                                                                                                                        & 1                                  \\
76                              & \stative                                     & they are speaking Portuguese in Brazil                                                                                                                     & 1                                  \\
76                              & \stative                                     & they speak Portuguese in Brazil                                                                                                                            & 0                                  \\
77                              & \stative                                     & \begin{tabular}[c]{@{}l@{}}and the production function is giving you the relationship\\ between input and output\end{tabular}                              & 1                                  \\
77                              & \stative                                     & \begin{tabular}[c]{@{}l@{}}and the production function gives you the relationship\\ between input and output\end{tabular}                                  & 0                                  \\
77                              & \article                                        & \begin{tabular}[c]{@{}l@{}}and the production function is giving you the relationship\\ between input and output\end{tabular}                              & 1                                  \\
77                              & \article                                        & \begin{tabular}[c]{@{}l@{}}and the production function gives you the relationship\\ between input and output\end{tabular}                                  & 1                                  \\
78                              & \stative                                     & he is having a television                                                                                                                                  & 1                                  \\
78                              & \stative                                     & he has a television                                                                                                                                        & 0                                  \\
79                              & \stative                                     & I think Nina must be knowing her sister                                                                                                                    & 1                                  \\
79                              & \stative                                     & I think Nina must know her sister                                                                                                                          & 0                                  \\
80                              & \stative                                     & we will be knowing how much the structure is getting deflected                                                                                             & 1                                  \\
80                              & \stative                                     & we will know how much the structure is getting deflected                                                                                                   & 0                                  \\
81                              & \extraarticle                 & Chandigarh was full of the employed people.                                                                                                                & 1                                  \\
81                              & \extraarticle                 & Chandigarh was full of employed people.                                                                                                                    & 0                                  \\
82                              & \extraarticle                 & She has a business experience.                                                                                                                             & 1                                  \\
82                              & \extraarticle                 & She has business experience.                                                                                                                               & 0                                  \\
83                              & \extraarticle                 & Because educated people get a good money.                                                                                                                  & 1                                  \\
83                              & \extraarticle                 & Because educated people get good money.                                                                                                                    & 0                                  \\
84                              & \extraarticle                 & They have a pressure from their in-laws.                                                                                                                   & 1                                  \\
84                              & \extraarticle                 & They have pressure from their in-laws.                                                                                                                     & 0                                  \\
85                              & \extraarticle                 & Here the life is busy.                                                                                                                                     & 1                                  \\
85                              & \extraarticle                 & Here life is busy.                                                                                                                                         & 0                                  \\
86                              & \andall                                        & \begin{tabular}[c]{@{}l@{}}So marketing keeps its communication with the different\\ embassies and all.\end{tabular}                                       & 1                                  \\
86                              & \andall                                        & \begin{tabular}[c]{@{}l@{}}So marketing keeps its communication with the different\\ embassies.\end{tabular}                                               & 0                                  \\
87                              & \andall                                        & \begin{tabular}[c]{@{}l@{}}Whereas we had lot of time and we didn’t have any TV\\ and all and we used to play outdoor games.\end{tabular}                  & 1                                  \\
87                              & \andall                                        & \begin{tabular}[c]{@{}l@{}}Whereas we had lot of time and we didn’t have any TV\\ and we used to play outdoor games.\end{tabular}                          & 0                                  \\
87                              & \andall                                        & \begin{tabular}[c]{@{}l@{}}Whereas we had lot of time and we didn’t have any TV\\ and all that stuff and we used to play outdoor games.\end{tabular}       & 0                                  \\
88                              & \andall                                        & So I did my schooling and all from there.                                                                                                                  & 1                                  \\
88                              & \andall                                        & So I did my schooling from there.                                                                                                                          & 0                                  \\
89                              & \andall                                        & \begin{tabular}[c]{@{}l@{}}We are like we are in touch, but not before when we\\ was in school and all.\end{tabular}                                       & 1                                  \\
89                              & \andall                                        & \begin{tabular}[c]{@{}l@{}}We are like we are in touch, but not before when we\\ was in school.\end{tabular}                                               & 0                                  \\
89                              & \agreement                                       & \begin{tabular}[c]{@{}l@{}}We are like we are in touch, but not before when we\\ was in school and all.\end{tabular}                                       & 1                                  \\
89                              & \agreement                                       & \begin{tabular}[c]{@{}l@{}}We are like we are in touch, but not before when we\\ was in school.\end{tabular}                                               & 1                                  \\
90                              & \andall                                        & \begin{tabular}[c]{@{}l@{}}My parents and siblings and all, they really enjoy\\ playing board games.\end{tabular}                                          & 1                                  \\
90                              & \andall                                        & \begin{tabular}[c]{@{}l@{}}My parents and siblings, they really enjoy playing\\ board games.\end{tabular}                                                  & 0                                  \\
90                              & \leftdis                                        & \begin{tabular}[c]{@{}l@{}}My parents and siblings and all, they really enjoy\\ playing board games.\end{tabular}                                          & 1                                  \\
90                              & \leftdis                                        & \begin{tabular}[c]{@{}l@{}}My parents and siblings, they really enjoy playing\\ board games.\end{tabular}                                                  & 1                                  \\
91                              & \copula                                          & I think she a teacher.                                                                                                                                     & 1                                  \\
91                              & \copula                                          & I think she is a teacher.                                                                                                                                  & 0                                  \\
92                              & \copula                                          & They all aggressive states.                                                                                                                                & 1                                  \\
92                              & \copula                                          & They are all aggressive states.                                                                                                                            & 0                                  \\
93                              & \copula                                          & Now they wearing American type of dresses.                                                                                                                 & 1                                  \\
93                              & \copula                                          & Now they are wearing American type of dresses.                                                                                                             & 0                                  \\
94                              & \copula                                          & So my parents from Gujarat.                                                                                                                                & 1                                  \\
94                              & \copula                                          & So my parents are from Gujarat.                                                                                                                            & 0                                  \\
95                              & \copula                                          & Sorry I can't come, everything busy in our life.                                                                                                           & 1                                  \\
95                              & \copula                                          & Sorry I can't come, everything is busy in our life.                                                                                                        & 0                                  \\
96                              & \resobj                               & The cake, I like it very much.                                                                                                                             & 1                                  \\
96                              & \resobj                               & I like the cake very much.                                                                                                                                 & 0                                  \\
96                              & \leftdis                                        & The cake, I like it very much.                                                                                                                             & 1                                  \\
97                              & \resobj                               & The book that I left it here, where is it?                                                                                                                 & 1                                  \\
97                              & \resobj                               & The book that I left here, where is it?                                                                                                                    & 0                                  \\
98                              & \resobj                               & My old life I want to spend it in India.                                                                                                                   & 1                                  \\
98                              & \resobj                               & My old life I want to spend in India.                                                                                                                      & 0                                  \\
98                              & \leftdis                                        & My old life I want to spend it in India.                                                                                                                   & 1                                  \\
99                              & \resobj                               & \begin{tabular}[c]{@{}l@{}}Some teachers when I was in school I liked them\\ very much.\end{tabular}                                                       & 1                                  \\
99                              & \resobj                               & \begin{tabular}[c]{@{}l@{}}Some teachers when I was in school I liked\\ very much.\end{tabular}                                                            & 0                                  \\
99                              & \leftdis                                        & \begin{tabular}[c]{@{}l@{}}Some teachers when I was in school I liked them\\ very much.\end{tabular}                                                       & 1                                  \\
100                             & \resobj                               & I’m going to find my bag which I left it in the room.                                                                                                      & 1                                  \\
100                             & \resobj                               & I’m going to find my bag which I left in the room.                                                                                                         & 0                                  \\
101                             & \ressubj                              & \begin{tabular}[c]{@{}l@{}}A person living in Calcutta, which he didn’t know Hindi\\ earlier, when he comes to Delhi he has to learn English.\end{tabular} & 1                                  \\
101                             & \ressubj                              & \begin{tabular}[c]{@{}l@{}}A person living in Calcutta, who didn’t know Hindi\\ earlier, when he comes to Delhi he has to learn English.\end{tabular}      & 0                                  \\
102                             & \ressubj                              & But now all kids they have a computer and all new technology.                                                                                              & 1                                  \\
102                             & \ressubj                              & But now all kids have a computer and all new technology.                                                                                                   & 0                                  \\
102                             & \leftdis                                        & But now all kids they have a computer and all new technology.                                                                                              & 1                                  \\
103                             & \ressubj                              & My daughter she is attending the University of Delhi.                                                                                                      & 1                                  \\
103                             & \ressubj                              & My daughter is attending the University of Delhi.                                                                                                          & 0                                  \\
103                             & \leftdis                                        & My daughter she is attending the University of Delhi.                                                                                                      & 1                                  \\
104                             & \ressubj                              & and that roommate, he will do an interview                                                                                                                 & 1                                  \\
104                             & \ressubj                              & and that roommate will do an interview                                                                                                                     & 0                                  \\
104                             & \leftdis                                        & and that roommate, he will do an interview                                                                                                                 & 1                                  \\
105                             & \ressubj                              & some people they are very nice                                                                                                                             & 1                                  \\
105                             & \ressubj                              & some people are very nice                                                                                                                                  & 0                                  \\
105                             & \leftdis                                        & some people they are very nice                                                                                                                             & 1                                  \\
106                             & \textsc{topicalized non-arg. const}                    & daytime I work for the courier service                                                                                                                     & 1                                  \\
106                             & \textsc{topicalized non-arg. const}                    & in the daytime I work for the courier service                                                                                                              & 1                                  \\
106                             & \textsc{topicalized non-arg. const}                    & I work for the courier service in the daytime                                                                                                              & 0                                  \\
106                             & \ppfronting & daytime I work for the courier service                                                                                                                     & 1                                  \\
107                             & \textsc{topicalized non-arg. const}                    & for many years I did not travel                                                                                                                            & 1                                  \\
107                             & \textsc{topicalized non-arg. const}                    & many years I did not travel                                                                                                                                & 1                                  \\
107                             & \textsc{topicalized non-arg. const}                    & I did not travel for many years                                                                                                                            & 0                                  \\
107                             & \ppfronting & many years I did not travel                                                                                                                                & 1                                  \\
108                             & \textsc{topicalized non-arg. const}                    & with your mother I love to go shopping                                                                                                                     & 1                                  \\
108                             & \textsc{topicalized non-arg. const}                    & I love to go shopping with your mother                                                                                                                     & 0                                  \\
109                             & \textsc{topicalized non-arg. const}                    & and in the background there are a lot of buildings                                                                                                         & 1                                  \\
109                             & \textsc{topicalized non-arg. const}                    & and there are a lot of buildings in the background                                                                                                         & 0                                  \\
110                             & \textsc{topicalized non-arg. const}                    & yeah, so my parent's house I go very often                                                                                                                 & 1                                  \\
110                             & \textsc{topicalized non-arg. const}                    & yeah, so to my parent's house I go very often                                                                                                              & 1                                  \\
110                             & \textsc{topicalized non-arg. const}                    & yeah, so I go very often to my parent's house                                                                                                              & 0                                  \\
110                             & \ppfronting & yeah, so my parent's house I go very often                                                                                                                 & 1                                 
\end{longtable}

\end{document}